\documentclass[11pt]{article}

\usepackage[preprint]{acl}

\usepackage{times}
\usepackage{latexsym}
\usepackage{amsmath}
\usepackage[T1]{fontenc}

\usepackage[utf8]{inputenc}

\usepackage{microtype}
\usepackage{tabularx}
\usepackage{inconsolata}

\usepackage{graphicx}
\usepackage{subcaption}
\usepackage[breakable,skins]{tcolorbox}
\usepackage{algorithm}
\usepackage{algorithmic}
\usepackage[table]{xcolor}
\definecolor{lightbluerow}{RGB}{242,247,255}
\definecolor{gainred}{RGB}{160,40,40}
\definecolor{dropblue}{RGB}{31,119,180}
\newcommand{\drop}[1]{\textcolor{dropblue}{{\scriptsize(#1)}}}
\usepackage{placeins}
\newcommand{\gain}[1]{{\scriptsize\textcolor{gainred}{(#1)}}}
\usepackage{enumitem}
\usepackage{booktabs} 
\usepackage{multirow} 
\tcbuselibrary{skins, breakable}

\usepackage{fvextra}                     
\usepackage{xcolor}
\usepackage{amssymb}                     

\newcommand{\exmatch}{%
  \textcolor{green!55!black}{\textbf{[\checkmark\ EX-Match]}}}
\newcommand{\exwrong}{%
  \textcolor{red!65!black}{\textbf{[\texttimes\ EX-Wrong]}}}

\fvset{
  breaklines=true,       
  breakanywhere=false,   
  breaksymbol={},        
  breakindent=1.5em,     
  fontsize=\small,
}

\newtcolorbox{promptbox}[1]{
  colback=blue!4!white, colframe=blue!65!black,
  title={\textbf{#1}}, fonttitle=\small\bfseries,
  breakable, left=5pt, right=5pt, top=3pt, bottom=3pt,
}
\newtcolorbox{modelbox}[1]{
  colback=green!4!white, colframe=green!55!black,
  title={\textbf{#1}}, fonttitle=\small\bfseries,
  breakable, left=5pt, right=5pt, top=3pt, bottom=3pt,
}
\newtcolorbox{goldbox}[1]{
  colback=orange!5!white, colframe=orange!70!black,
  title={\textbf{#1}}, fonttitle=\small\bfseries,
  breakable, left=5pt, right=5pt, top=3pt, bottom=3pt,
}
\newtcolorbox{feedbackbox}[1]{
  colback=red!4!white, colframe=red!60!black,
  title={\textbf{#1}}, fonttitle=\small\bfseries,
  breakable, left=5pt, right=5pt, top=3pt, bottom=3pt,
}

\title{Progress-SQL: Improving Reinforcement Learning for Text-to-SQL via Progressive Rewards}

\author{
Shihao Zhang,
Xiaoman Wang,
Yuan Liu,
Yunshi Lan\thanks{Corresponding author.},
Weining Qian
\\
East China Normal University
\\
\texttt{shzhang@stu.ecnu.edu.cn},
\texttt{yslan@dase.ecnu.edu.cn}
}

\begin{document}
\maketitle
\begin{abstract}
Reinforcement learning has recently shown promise in improving large language models for Text-to-SQL generation, yet existing methods typically optimize one-shot rewards defined over a single SQL state. Such rewards provide limited guidance for iterative SQL correction and are insufficient to capture the improvement of multi-turn SQL refinement. In this paper, we propose Progress-SQL, a multi-turn reinforcement learning framework with progressive rewards for Text-to-SQL. Our approach introduces an Oracle-guided Diagnostic Tree (ODT), which abstracts SQL queries into clause-level structural profiles and produces diagnostic feedback for next-turn refinement. To provide dense and robust reward signals, we combine ODT-based structural alignment with lexical alignment and define a progressive reward that measures the improvement from the initial SQL to the final SQL. We further incorporate a progression latency reward that favors earlier correctness and an execution status reward that encourages recovery from the invalid SQL. Experiments on BIRD, Spider, and Spider robustness variants demonstrate that our method consistently improves Text-to-SQL performance across both primary and robustness evaluations. Our code is released at \url{https://github.com/YooYoo67/ProgressSQL}.
\end{abstract}

\section{Introduction}
\label{sec:intro}

Large Language Models (LLMs) have significantly advanced Text-to-SQL parsing~\citep{li2024codes,DBLP:journals/pvldb/LiWZHZJWZCSCL25}.
Recent Reinforcement Learning (RL) methods further improve these models by optimizing one-shot rewards~\citep{pourreza2025reasoning,zhang2025rewardsqlboostingtexttosqlstepwise,DBLP:conf/emnlp/WengWCZLSZYZHYSC25,ma2026sql}, where the single turn rollout is measured by a one-shot reward.
This reward is computed solely based on the execution result of the SQL.
However, such a reward is often sparse, which provides limited guidance on SQL generation.
This results in inefficient optimization for the exploration of correct SQLs, especially for complex SQLs involving  joins and aggregation~\cite{pourreza2025reasoning}.

\begin{figure}[t]
  \centering
  \includegraphics[width=1.0\columnwidth]{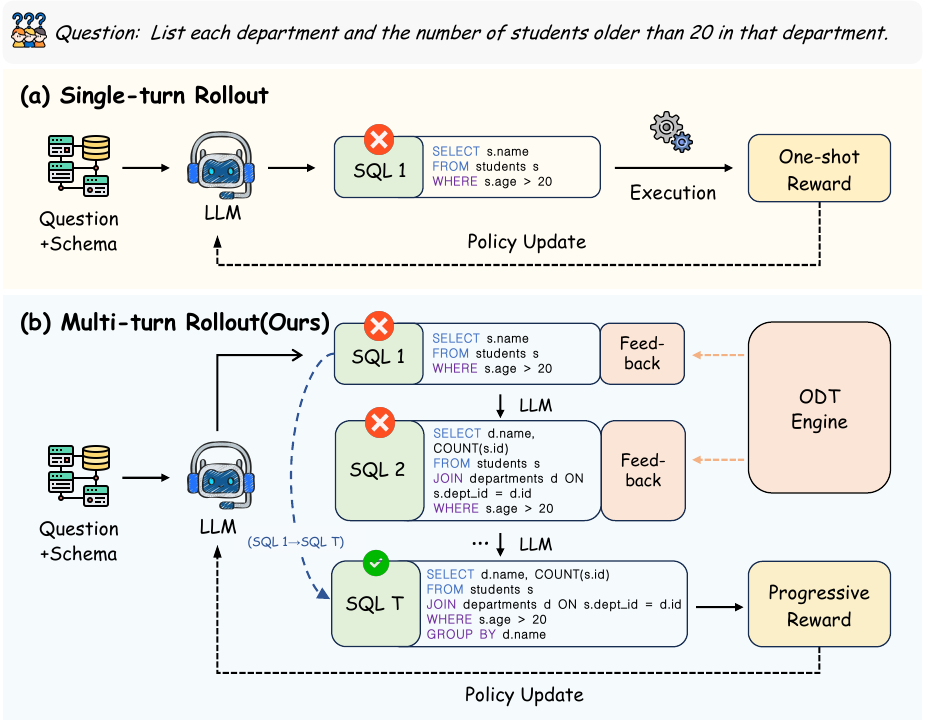}
  \caption{Comparison of reward paradigms.
      \textbf{(a) Single-turn Rollout}: the policy model generates a single SQL and receives
      a reward signal after execution.
      \textbf{(b) Multi-turn Rollout with Progressive Reward (Ours)}: the policy model iteratively refines its SQL
      over $T$ turns guided by ODT engine. The progressive reward measures improvement from
      the first SQL to the final SQL.}
  \label{fig:reward_comparison}
\end{figure}

A cutting-edge study, SkyRL-SQL~\cite{liu2025skyrlsql}, introduced multi-turn rollout for the Text-to-SQL task, where multi-turn interaction is conducted between the LLMs and engines. The LLM generates a sequence of SQLs for a question.
For each turn, the LLM obtains the execution result from the engine and takes it into consideration for the next-turn generation.
Nevertheless, SkyRL-SQL collects the reward based on the last-turn generation, which cannot yet break away from the limitation of the one-shot reward.
In other words, \textbf{one-shot reward is not enough to capture the dynamic behavior of multi-turn rollout}.

To address this limitation, we propose Progress-SQL, a multi-turn RL framework with progressive rewards for Text-to-SQL, as shown in
Figure~\ref{fig:reward_comparison}.
Specifically, we first introduce an Oracle-guided Diagnostic Tree (ODT), which
abstracts SQL queries into clause-level structural profiles and generates
diagnostic feedback for next-turn refinement.
By comparing the predicted ODT with the gold ODT during training, the model can
revise its SQL prediction according to the structured feedback. Unlike one-shot rewards, our progressive reward is defined over the SQL
trajectory and measures whether the final SQL improves over the initial SQL in
terms of structural and lexical alignment.
Together with a progression latency reward and an execution status reward, the
objective favors trajectories that improve effectively, reach correctness
earlier, and recover from invalid SQL predictions.
We evaluate our RL method on widely used Text-to-SQL benchmarks, including BIRD and Spider.
Based on the 7B backbone, our method improves the base model by an average of 8.5\% in execution accuracy across BIRD Dev, Spider Dev, and Spider Test, and by 6.3\% in test-suite accuracy on Spider Dev.
Compared with LLMs tuned by existing RL methods, our method achieves competitive or superior performance after fine-tuning.

Our contributions are summarized as follows:
\begin{itemize}[leftmargin=*]
    \item We propose Progress-SQL, a multi-turn RL framework for Text-to-SQL. By defining an ODT for clause-level SQL diagnosis, Progress-SQL collects fine-grained feedback for next-turn SQL generation.
    \item We design a progressive reward that explicitly measures the improvement from the initial SQL to the final SQL, complemented by early-correctness and execution-status rewards for efficient and robust refinement.
    \item Extensive experiments on multiple Text-to-SQL benchmarks demonstrate that our method consistently improves both execution accuracy and test-suite accuracy with different base models.
\end{itemize}

\section{Related Work}
\label{sec:related_work}

\subsection{Text-to-SQL with Large Language Models}

Due to the outstanding performance of LLMs in various NLP tasks, we observe a methodology trend of Text-to-SQL systems from heuristic rules and deep learning to LLMs~\cite{zelle1996learning,popescu2003towards,li2014constructing,yu2018typesql,wang2020rat}.
In the early stages, LLM-based Text-to-SQL systems relied heavily on In-Context Learning (ICL) and structured prompting strategies. DIN-SQL~\citep{DBLP:conf/nips/PourrezaR23} utilizes prompting instruction to decompose complex queries into sub-problems, while DAIL-SQL~\citep{DBLP:journals/pvldb/GaoWLSQDZ24} constructs efficient few-shot demonstrations via question skeleton matching. More recent pipelines such as CHASE-SQL~\citep{DBLP:conf/iclr/PourrezaL0CTKGS25} and XiYan-SQL~\citep{DBLP:journals/tkde/LiuZGLLSHGLDZ26} further combine candidate generation and selection strategies to refresh the accuracy on competitive benchmarks. Alongside prompting, post-training has become a standard paradigm~\citep{li2024codes,DBLP:journals/pvldb/LiWZHZJWZCSCL25}, improving the Text-to-SQL performance for open-source models in curated training splits.

\subsection{Reinforcement Learning for Text-to-SQL}

Regarding the post-training procedure, recent studies have framed Text-to-SQL as a sequential decision-making problem optimized with reinforcement learning, where reward design is crucial to measuring SQL quality. Recent studies have explored RL for Text-to-SQL by exploring different rewards. The most direct signal is binary execution accuracy (EX), but it is highly sparse because rewards are obtained only when the generated SQL matches the gold execution result. To provide denser supervision, Reward-SQL~\citep{zhang2025rewardsqlboostingtexttosqlstepwise} introduces a Process Reward Model for step-wise reasoning supervision, while Reasoning-SQL~\citep{pourreza2025reasoning} designs SQL-specific partial rewards such as schema-linking accuracy, n-gram similarity, and syntax validity. Graph-Reward-SQL~\citep{DBLP:conf/emnlp/WengWCZLSZYZHYSC25} further incorporates structural tree matching to capture logical alignment. SkyRL-SQL~\citep{liu2025skyrlsql} extends RL to multi-turn SQL refinement, but its reward is still derived from the final SQL state. However, these methods mainly define rewards over single or final SQL states, leaving trajectory-level revision behavior underexplored.
However, these methods mainly define rewards over single or final SQL states, leaving the trajectory-level revision behavior underexplored.

\section{Preliminaries}

\subsection{Problem Definition}
The Text-to-SQL task can be formulated as a semantic parsing problem that translates a natural language question into an executable SQL query. Formally, let $q = \{q_1, q_2, \dots, q_{l}\}$ denote the natural language question and $S$ denote the corresponding database schema (comprising tables, columns, and foreign key constraints). Given the input context $x = (q, S)$, the goal is to generate a target SQL query $y = \{w_1, w_2, \dots, w_{m}\}$ which consists of a sequence of tokens and can correctly retrieve the answers from the database engine $\mathcal{E}$. 


\subsection{Reinforcement Learning Protocol}

Existing studies~\citep{schulman2017proximal} view the generation process through a parameterized policy $\pi_\theta(y\mid x)$, which is initialized from a pre-trained instruction-tuned LLM. We denote it as the reference policy $\pi_{\text{ref}}$.
The Text-to-SQL task can be solved via the reinforcement learning protocol, which optimize the policy model by maximizing:
\begin{align*}
\mathcal{J}(\theta) = & \mathbb{E}_{x \sim P(x), y \sim \pi_{\theta}} [\mathcal{R}(y, y^*) \\
& - \beta \mathbb{D}_{\text{KL}}(\pi_{\theta}(\cdot | x) \| \pi_{\text{ref}} (\cdot | x)) ],
\end{align*}
where $x$ is sampled from a Text-to-SQL dataset, $\pi_{\text{ref}}$ is the initial reference model, and $\pi_{\theta}$ is iteratively updated.
KL divergence penalty prevents the updated policy from degrading its fundamental language capabilities during training.
Reward function measures the distance between the generated SQL query $y$ and the gold SQL query $y^*$.
The similar protocol with different RL algorithms such as GRPO~\citep{shao2024deepseekmathpushinglimitsmathematical}, GSPO~\citep{zheng2025groupsequencepolicyoptimization} are also widely utilized to solve the task.

\subsection{One-shot Reward Design}


\label{subsec:oneshot_reward}

In RL protocol, one of the core objectives is to maximize the reward $\mathcal{R}(y, y^*)$, which can be deemed as a one-shot measurement between the generated SQL and the gold SQL.
We notice that a number of one-shot rewards have been proposed for solving Text-to-SQL task, which can be summarized below.


\noindent \textbf{Execution Matching}. Measuring the consistency of the execution results between the generated SQL and the gold SQL is an intuitive way to evaluate the policy model.
Standard approach formulates it as a binary reward ($1$ for exact match, $0$ otherwise)~\cite{pourreza2025reasoning}. 
Due to the sparse reward signal, recent studies improve the reward by introducing fractional execution based on the proportion of matching columns and cells~\cite{hao2025paverl,papicchio2025think2sql}.




\noindent \textbf{Query Matching}.
To construct a denser reward landscape, researchers incorporate static structural similarities between $y$ and $y^*$. Reasoning-SQL \cite{pourreza2025reasoning} calculates the Jaccard similarity of extracted schema items and 2-grams between the generated and the annotated SQLs, thus guiding the query alignment beyond database execution. Besides the semantics matching, syntax is also considered in the reward~\cite{pourreza2025reasoning,ali2025state}.

\noindent \textbf{Format and Process Regularization}.
With the advent of reasoning models (\emph{e.g.}, DeepSeek-R1~\citep{Guo_2025}), enforcing specific Chain-of-Thought (CoT) behaviors has become integral to reward design. Models are incentivized via format rewards to encapsulate their reasoning processes within specific tags (\emph{e.g.}, \texttt{<think>} and \texttt{<sql>}) ~\citep{ma2026sql, papicchio2025think2sql, pourreza2025reasoning}. Furthermore, to prevent reward hacking and excessive verbosity, regularization terms such as tag count \citep{papicchio2025think2sql} and length penalties ~\citep{ma2026sql} are applied to penalize redundant reasoning. And some studies even involve schema keywords, runtime log in the rewards~\citep{berdnyk2025llm,ma2026sql}.


\section{Methodology}
\begin{figure*}[t] 
    \centering
    \includegraphics[width=0.95\textwidth]{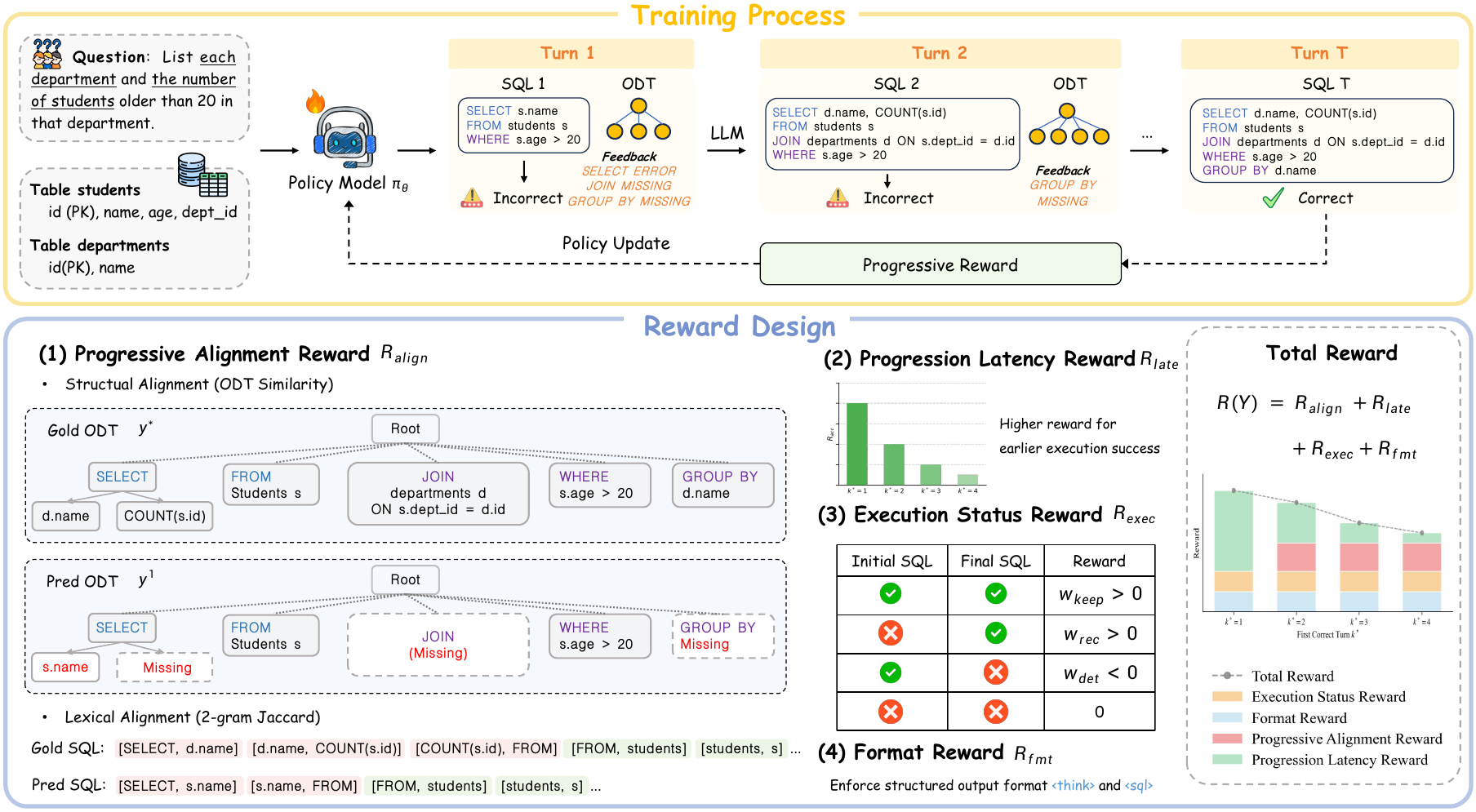} 
    \vspace{-0.2cm}
    \caption{Overall framework of Progress-SQL, our multi-turn reinforcement learning method for Text-to-SQL.
The policy model iteratively generates SQL queries and receives ODT-based diagnostic feedback after each execution.
The final trajectory is optimized using progressive rewards that combine structural/lexical alignment improvement, progression latency reward, execution-status transition reward, and format reward.}
    \label{fig:main_arch}
\end{figure*}

\subsection{Multi-turn Rollout with ODT Feedback}
\label{subsec:multi_turn_rollout}

Standard RL methods for Text-to-SQL usually implement single-turn rollout, where the policy generates a SQL query once and receives sparse rewards from the single decoded SQL.
To address this limitation, we extend the standard rollout process into a multi-turn SQL debugging trajectory. Inspired by  SkyRL-SQL~\citep{liu2025skyrlsql}, we allow the policy model to iteratively revise its SQL prediction based on feedback from previous turns as shown in Figure~\ref{fig:main_arch}. 
Remarkably, we introduce an
\textit{Oracle-guided Diagnostic Tree} (ODT) as the structural feedback after each attempt to refine the next-turn rollout.

Formally, given a user question $q$ and database schema $S$, the initial input is
defined as:
\begin{align*}
    x^{(1)} = (q, S).
\end{align*}
At the $t$-th turn, the policy model $\pi_\theta$ generates a new SQL prediction as the rollout with the consideration of the input $x^{(t)}$:
\begin{align*}
    y^{(t)} \sim \pi_\theta(\cdot \mid x^{(t)}).
\end{align*}

Regarding $x^{(t)}$, we construct the input by appending ODT structural
feedback generated from the previous prediction:
\begin{align*}
    f^{(t-1)} &= \mathrm{ODT}(y^{(t-1)}, y^*), \\
    x^{(t)} &= (q, S, y^{(1)}, f^{(1)}, \dots, y^{(t-1)}, f^{(t-1)}).
\end{align*}
The rollout continues until the model produces a correct executable SQL or the
maximum interaction budget $T$ is reached.
This process yields a multi-turn
trajectory $Y = (y^{(1)}, y^{(2)}, \dots, y^{(T)})$, which is then evaluated by our progressive rewards.

\noindent \textbf{Oracle-guided Diagnostic Tree}. Above, directly comparing a predicted SQL with the gold SQL is non-trivial due to the compositional and nested syntax of SQL queries.
Following the common practice of using Abstract Syntax Trees (ASTs) to capture SQL syntactic structures, we parse both the predicted SQL $y$ and the gold SQL $y^*$ into ASTs.
Rather than directly matching raw AST nodes, we abstract each SQL into an Oracle-guided Diagnostic Tree (ODT).
As shown in Figure~\ref{fig:main_arch}, each node in the tree represents a clause-level structural profile, such as selected columns, join signatures, filtering predicates, grouping columns, ordering clauses, or nested subqueries.
By comparing the ODTs of $y$ and $y^*$, the diagnostic module produces two outputs: a structural similarity score and a set of clause-level error tags.
The error tags are verbalized into diagnostic feedback for the next-turn refinement, while the structural similarity score serves as the structural alignment term in our reward function (refer to Section~\ref{subsec:reward_formulation}).
For example, in Figure~\ref{fig:main_arch}, the first-turn SQL only selects
student names with an age predicate, while the gold SQL requires department-level
aggregation by joining \texttt{students} with \texttt{departments} and grouping
by department. The ODT comparison detects mismatches in selection, join, and
grouping structures, producing tags such as \texttt{SELECT\_ERROR},
\texttt{JOIN\_MISSING}, and \texttt{GROUP\_BY\_MISSING}. These tags are verbalized
as feedback to guide the next-turn SQL refinement.

ODT acts as a fixed, non-differentiable diagnostic component of the environment.
It compares the predicted SQL with the gold SQL and appends discrete error tags as next-turn observations; these tags do not participate in back-propagation, and gradients are computed only through the policy model over generated tokens. Details of the multi-turn algorithm and ODT construction are provided in Appendix~\ref{app:rollout_algorithm} and Appendix~\ref{app:odt_scoring}, respectively.

\subsection{Progressive Reward Formulation}
\label{subsec:reward_formulation}

Given a multi-turn SQL trajectory
$Y=(y^{(1)},y^{(2)},\dots,y^{(T)})$ and the gold SQL $y^*$, our goal is not only
to evaluate the final prediction, but also to measure whether the trajectory
makes meaningful progress. We therefore design a comprehensive reward composed
    of four parts: \textit{Progressive Alignment Reward}, \textit{Progression Latency Reward}, \textit{Execution Status Reward}, and \textit{Format Reward}.

\paragraph{Progressive Alignment Reward.}
To measure the similarity between the predicted and ground truth SQLs, we define both structural and lexical aspects of alignment:
\begin{itemize}[leftmargin=*]
    \item \textbf{Structural Alignment.} The structural alignment score is computed by the ODT tree described above.
Specifically, we abstract both the generated SQL and the gold SQL into ODTs and compare their clause-level structural profiles.
To focus on structural correspondence, lexical values are normalized or replaced with placeholders before matching.
    For each node, we compute a similarity score by combining local clause similarity and child-subtree similarity:
\begin{equation*}
    s_{\mathrm{node}} =
    \alpha \cdot s_{\mathrm{local}}
    + (1-\alpha) \cdot s_{\mathrm{child}},
\end{equation*}
where $s_{\mathrm{local}}$ measures the weighted Jaccard similarity over
clause-level feature sets. 
For nested structures, $s_{\mathrm{child}}$ is computed recursively by matching
child subtrees under node-type compatibility constraints. The final structural
alignment score is given by the root node score:
\begin{align*}
    \mathcal{F}_{\mathrm{struct}}(y,y^*) = s_{\mathrm{root}}.
\end{align*}
Hence, structural alignment measures clause-level structural consistency between $y$ and $y^*$ while reducing sensitivity to surface lexical differences. The detailed ODT construction and scoring are provided in Appendix~\ref{app:odt_scoring}.
\item \textbf{Lexical Alignment.}
Structural parsing may fail when the generated SQL contains severe syntax
errors, especially during early training. To avoid zero-reward regions in such
cases, we further introduce a lexical fallback score. Specifically, we compute
the 2-gram Jaccard similarity between the tokenized generated SQL and the gold
SQL, denoted as $\mathcal{F}_{\mathrm{lex}}(y,y^*)$.
The lexical score provides a dense signal when structural parsing is
unavailable.
\end{itemize}

Rather than rewarding only endpoint correctness, we explicitly measure whether
the final SQL improves over the initial prediction.
The final alignment score is defined as:
\begin{align*}
    \mathcal{F}(y,y^*) =&
    \frac{1}{2}
    \left(
    \mathcal{F}_{\mathrm{struct}}(y,y^*) +
    \mathcal{F}_{\mathrm{lex}}(y,y^*)
    \right) , \\
    \Delta = &
    \mathcal{F}(y^{(T)}, y^*)
    -
    \mathcal{F}(y^{(1)}, y^*).
\end{align*}
This endpoint-difference design reduces the influence of intermediate
oscillations and encourages the trajectory to end with a better SQL query than
it starts with.

The progressive reward is defined as:
\begin{equation}
    \mathcal{R}_{\mathrm{align}} =
    \begin{cases}
        \omega_{\mathrm{align}}^{+} \cdot \Delta,
        & \text{if } \Delta > 0, \\
        \omega_{\mathrm{align}}^{-},
        & \text{if } \Delta \leq 0, \\
    \end{cases}
\end{equation}
where $\omega_{\mathrm{align}}^{+}$ controls the reward for positive progress and
$\omega_{\mathrm{align}}^{-}$ penalizes stagnant trajectories. This encourages the model to perform meaningful revisions instead of meaningless repeats.

\paragraph{Progression Latency Reward.}
To encourage immediate progression to SQL correction, we terminate the rollout once the generated SQL first matches the execution result of the gold SQL. Let $k^*$ denote the first successful turn, where success means execution result matches with the gold SQL. We apply a geometric decay to late corrections:
\begin{equation}
    \mathcal{R}_{\mathrm{late}} =
    \begin{cases}
        \omega_{\mathrm{acc}} \cdot \gamma^{k^*-1}, & 1 \le k^* \le T, \\
        0, & \text{otherwise},
    \end{cases}
\end{equation}
where $\omega_{\mathrm{acc}}>0$ controls the base reward for execution correctness and
$\gamma \in (0,1)$ is the per-turn decay factor.
Earlier successful turns therefore receive larger rewards, encouraging the model to produce a correct SQL faster.

\paragraph{Execution Status Reward.}
Execution correctness can be too sparse, but executability itself provides useful
information about whether the model is recovering from invalid SQL. We therefore
reward status transitions in executability between the initial and final predictions.
Let $\mathcal{E}(y)$ be a boolean indicator that returns true if $y$ can be
executed by the database engine without syntax or runtime errors. We define:
\begin{equation}
    \mathcal{R}_{\mathrm{exec}} =
    \begin{cases}
        \omega_{\mathrm{keep}},
        & \text{if } \mathcal{E}(y^{(1)}) \land \mathcal{E}(y^{(T)}), \\
        \omega_{\mathrm{rec}},
        & \text{if } \neg \mathcal{E}(y^{(1)}) \land \mathcal{E}(y^{(T)}), \\
        \omega_{\mathrm{det}},
        & \text{if } \mathcal{E}(y^{(1)}) \land \neg \mathcal{E}(y^{(T)}), \\
        0,
        & \text{otherwise}.
    \end{cases}
\end{equation}
Here, $\omega_{\mathrm{keep}}>0$ rewards SQLs that remain executable across refinement,
$\omega_{\mathrm{rec}}>0$ rewards recovery from an initially unexecutable SQL to an executable one,
and $\omega_{\mathrm{det}}<0$ penalizes deterioration from an executable SQL to an invalid one.

\paragraph{Format Reward.}
Following prior endpoint-reward designs~\citep{ma2026sql, papicchio2025think2sql, pourreza2025reasoning}, 
we define the format reward as 
$\mathcal{R}_{\mathrm{fmt}}=\omega_{\mathrm{fmt}}$ if the final output follows 
the required \texttt{<think>} and \texttt{<sql>} templates, and 
$\mathcal{R}_{\mathrm{fmt}}=0$ otherwise.

\paragraph{Overall Reward.}
The final trajectory-level reward is the sum of all components:
\begin{equation}
    \mathcal{R}(Y) =
    \mathcal{R}_{\mathrm{fmt}}
    + \mathcal{R}_{\mathrm{exec}}
    + \mathcal{R}_{\mathrm{late}}
    + \mathcal{R}_{\mathrm{align}}.
\end{equation}
This reward jointly evaluates format validity, executability transition,
endpoint correctness, and the improvement of progressive alignment over the multi-turn
trajectory.

\noindent \textbf{Reward Analysis}. Figure~\ref{fig:main_arch} illustrates the theoretical upper bound of reward under different first correct turn $k^*$ with keeping format and execution status unchanged.
We elaborately configure the defined weights (refer to Appendix~\ref{app:implementation_details}) to show an overall downward trend as shown in the Figure~\ref{fig:main_arch}.
This upper-bound trend shows that the reward design explicitly favors earlier positive progression and also encourages the success of the earlier rollout.

\section{Experiments}
\label{sec:experimental_setup}

\subsection{Experimental Setup}

\noindent \textbf{Datasets}.
\label{subsec:datasets}
We conduct experiments on widely recognized Text-to-SQL benchmarks:
\begin{itemize}[leftmargin=*]
    \item \textbf{BIRD} \citep{DBLP:conf/nips/LiHQYLLWQGHZ0LC23}: A large-scale, cross-domain dataset featuring complex reasoning questions and real-world database schema. It is highly challenging and serves as our primary testbed for multi-turn reasoning capabilities.
    \item \textbf{Spider \& its variants}: We also evaluate on the standard \textbf{Spider} \citep{DBLP:conf/emnlp/YuZYYWLMLYRZR18} dataset to assess cross-domain generalization. To further test the model's robustness against schema synonyms and domain knowledge perturbations, we report results on its challenging variants: \textbf{Spider-Syn} \citep{gan-etal-2021-towards}, \textbf{Spider-Realistic} \citep{DBLP:conf/naacl/DengAMPSR21}, and \textbf{Spider-DK} \citep{DBLP:conf/emnlp/GanCP21}.
\end{itemize}

\begin{table*}[t]
  \centering
  \small
  \setlength{\tabcolsep}{4.5pt}
  \renewcommand{\arraystretch}{0.92}

  \begin{tabular*}{\textwidth}{@{\extracolsep{\fill}}lccccc@{}}
    \toprule
    \multirow{2}{*}{\textbf{Model}} 
    & \multirow{2}{*}{\textbf{Reward}} 
    & \textbf{BIRD Dev} 
    & \multicolumn{2}{c}{\textbf{Spider Dev}} 
    & \textbf{Spider Test} \\
    \cmidrule(lr){3-3} \cmidrule(lr){4-5} \cmidrule(l){6-6}
    & & \textbf{EX(\%)} & \textbf{EX(\%)} & \textbf{TS(\%)} & \textbf{EX(\%)} \\
    \midrule

    \multicolumn{6}{@{}l}{\textit{Base and Supervised Fine-tuned Models}} \\
    Qwen2.5-Coder-7B-Instruct & -- & 52.3 & 79.7 & 73.8 & 81.8 \\
    Qwen2.5-Coder-14B-Instruct & -- & 60.2 & 83.4 & 77.0 & 86.1 \\
    OmniSQL-7B \citep{DBLP:journals/pvldb/LiWZHZJWZCSCL25} & -- & 63.0 & 85.8 & 81.0 & 87.2 \\
    SFT CodeS-7B \citep{li2024codes} & -- & 57.2 & 85.4 & 80.3 & -- \\
    \midrule

    \multicolumn{6}{@{}l}{\textit{Other Post-training Methods}} \\
    Reasoning-SQL-7B \citep{pourreza2025reasoning} & One-shot & 64.0 & 78.7 & -- & -- \\
    Reward-SQL-7B \citep{zhang2025rewardsqlboostingtexttosqlstepwise} & One-shot & 59.7 & 77.1 & -- & -- \\
    Graph-Reward-SQL-7B \citep{DBLP:conf/emnlp/WengWCZLSZYZHYSC25} & One-shot & 63.0 & 81.6 & -- & -- \\
    SQL-R1-7B$^\dagger$ \citep{ma2026sql} & One-shot & 59.8 & 81.8 & 67.6 & 81.7 \\
    SkyRL-SQL-7B \citep{liu2025skyrlsql} & One-shot & -- & 83.9 & -- & 85.2 \\
    SQL-Trail-7B$^\dagger$ \citep{hua2026sql} & One-shot & 57.0 & 85.6 & 77.4 & 84.0 \\
    \midrule

    \multicolumn{6}{@{}l}{\textit{Our Reinforcement Learning Models}} \\
    Progress-SQL-7B(Ours) & Progressive 
    & 64.3\gain{+12.0} 
    & 87.1\gain{+7.4} 
    & 80.9\gain{+7.1} 
    & 87.8\gain{+6.0} \\
    Progress-SQL-14B(Ours) & Progressive 
    & 67.5\gain{+7.3} 
    & 87.3\gain{+3.9} 
    & 80.4\gain{+3.4} 
    & 88.1\gain{+2.0} \\
    \bottomrule
  \end{tabular*}
\caption{
  Performance comparison on the primary Text-to-SQL benchmarks.
  All comparative 7B baselines post-trained by RL algorithm are based on Qwen2.5-Coder-7B-Instruct.
  We report EX and TS if they are available.
  Numbers in red fonts indicate absolute gains over the same-scale base model.
  $^\dagger$ indicates results reproduced by us following the data construction
described in the corresponding papers and without additional task-specific
supervised fine-tuning.
  }
  \label{tab:main_results}
\end{table*}

\noindent \textbf{Evaluation Metrics}.
\label{subsec:metrics}
We employ two standard automatic evaluation metrics in recent Text-to-SQL research:
\begin{itemize}[leftmargin=*]
    \item \textbf{Execution accuracy (EX)}: Measures whether the execution output of the generated SQL matches the exact result of the ground-truth SQL on the target database.
    \item \textbf{Test-Suite accuracy (TS)}: A more rigorous metric that evaluates the generated SQL across multiple augmented database test cases to ensure generalization and prevent false positives caused by coincidental execution matches.
\end{itemize}

\noindent \textbf{Baselines}.
\label{subsec:baselines}
We compare our models against two categories of strong baselines, corresponding to the groupings in Table~\ref{tab:main_results}.

\begin{itemize}[leftmargin=10pt]

    \item \textbf{Base and Supervised Fine-tuned Models}: This category includes
    our backbone models, Qwen2.5-Coder-7B/14B-Instruct, evaluated before
    applying our RL training, as well as supervised fine-tuned Text-to-SQL models
    such as OmniSQL~\citep{DBLP:journals/pvldb/LiWZHZJWZCSCL25} and
    SFT CodeS~\citep{li2024codes}. This comparison measures the gains brought
    by our RL framework beyond standard supervised or instruction-tuned models.

    \item \textbf{Comparative RL methods}: This category covers recent
    Text-to-SQL systems trained with reinforcement learning or reasoning-oriented
    post-training objectives, including both single-turn and multi-turn RL methods.
    All models in this group are based on Qwen2.5-Coder-7B-Instruct, making them
    directly comparable to our 7B model under the same backbone family. Specifically,
    we include Reasoning-SQL~\citep{pourreza2025reasoning},
    Reward-SQL~\citep{zhang2025rewardsqlboostingtexttosqlstepwise},
    Graph-Reward-SQL~\citep{DBLP:conf/emnlp/WengWCZLSZYZHYSC25},
    SQL-R1~\citep{ma2026sql}, SkyRL-SQL~\citep{liu2025skyrlsql}, and
    SQL-Trail~\citep{hua2026sql}. The first four methods use single-turn rollouts
    with one-shot rewards, while SkyRL-SQL and SQL-Trail introduce multi-turn
    refinement but still optimize rewards mainly associated with final SQL states.
    These baselines provide direct comparisons to our progressive multi-turn RL
    framework.
\end{itemize}

\noindent \textbf{Implementation Details}.
\label{subsec:implementation_details}
We adopt Qwen2.5-Coder-7B/14B-Instruct~\citep{hui2024qwen2} as base models and train them with GRPO~\citep{shao2024deepseekmathpushinglimitsmathematical} using the \texttt{verl} framework~\citep{sheng2024hybridflow}.
Following prior multi-turn Text-to-SQL training setups~\citep{hua2026sql}, we sample $G=8$ rollouts per input with temperature $\tau=1.0$.
Method-specific hyperparameters are selected by small-scale pilot experiments on the development set and then kept fixed across all experiments, with the maximum interaction turn set to $T=4$ and the per-turn decay factor set to $\gamma=0.5$.
ODT is used only during training for feedback construction and progressive reward computation; evaluation follows standard single-turn inference without ODT feedback, with transfer behavior analyzed in Appendix~\ref{app:single_turn_inference}.
Test results are reported with majority voting over 8 samples (Vote@8).
Full hyperparameters are provided in Appendix~\ref{app:implementation_details}.

\subsection{Main Results}
\label{subsec:main_results}

Table~\ref{tab:main_results} reports the performance on the widely recognized Text-to-SQL benchmarks. 
We have the following observations accordingly:

(1) LLMs fine-tuned by our RL method consistently improve over their corresponding base models, which indicates the generalization of our RL method. 
For the 7B backbone LLMs, Progress-SQL achieves an average EX improvement of 
$8.5\%$ across the three benchmarks, together with a $7.1\%$
TS improvement on Spider Dev.
For the 14B backbone LLMs, Progress-SQL achieves an average EX improvement of 
$4.4\%$ across the three benchmarks, together with a $3.4\%$
TS improvement on Spider Dev.
(2) Compared with existing post-training methods, our approach achieves competitive or superior performance across the evaluated benchmarks. In particular, Progress-SQL-7B reaches $87.1\%$ EX on Spider Dev and $87.8\%$ EX on Spider Test, surpassing recent 7B SQL RL methods (e.g. SQL-R1-7B and SkyRL-SQL-7B) on these benchmarks. Progress-SQL-14B further improves the results, suggesting that our progressive rewards can scale to stronger base models.

\label{app:robustness evaluation}
\begin{table*}[t]
  \centering
  \small
  \setlength{\tabcolsep}{4.2pt}
  \renewcommand{\arraystretch}{0.92}

  \begin{tabular*}{\textwidth}{@{\extracolsep{\fill}}lcccccc@{}}
    \toprule
    \multirow{2}{*}{\textbf{Model}} 
    & \multirow{2}{*}{\textbf{Reward}} 
    & \multicolumn{2}{c}{\textbf{Spider-Syn}} 
    & \multicolumn{2}{c}{\textbf{Spider-Realistic}} 
    & \textbf{Spider-DK} \\
    \cmidrule(lr){3-4} \cmidrule(lr){5-6} \cmidrule(l){7-7}
    & & \textbf{EX(\%)} & \textbf{TS(\%)} & \textbf{EX(\%)} & \textbf{TS(\%)} & \textbf{EX(\%)} \\
    \midrule

    \multicolumn{7}{@{}l}{\textit{Base and Supervised Fine-tuned Models}} \\
    Qwen2.5-Coder-7B-Instruct & -- & 70.9 & 62.3 & 74.4 & 66.9 & 66.9 \\
    Qwen2.5-Coder-14B-Instruct & -- & 72.3 & 65.6 & 81.5 & 74.6 & 72.3 \\
    OmniSQL-7B \citep{DBLP:journals/pvldb/LiWZHZJWZCSCL25} & -- & 77.0 & 70.2 & 80.5 & 75.0 & 74.6 \\
    SFT CodeS-7B \citep{li2024codes} & -- & 76.9 & 70.0 & 82.9 & 77.2 & 72.0 \\
    \midrule

    \multicolumn{7}{@{}l}{\textit{Other Post-training Methods}} \\
    Reasoning-SQL-7B \citep{pourreza2025reasoning} & One-shot & 69.3 & -- & -- & -- & 73.3 \\
    SQL-R1-7B$^\dagger$ \citep{ma2026sql} & One-shot & 73.7 & 58.0 & 80.1 & 61.6 & 69.9 \\
    SkyRL-SQL-7B \citep{liu2025skyrlsql} & One-shot & 73.7 & -- & 81.1 & -- & 72.0 \\
    SQL-Trail-7B$^\dagger$ \citep{hua2026sql} & One-shot & 73.7 & 65.1 & 83.7 & 74.4 & 75.4 \\
    \midrule

    \multicolumn{7}{@{}l}{\textit{Our Reinforcement Learning Models}} \\
    Progress-SQL-7B(Ours) & Progressive 
    & 76.6 \gain{+5.7}
    & 68.8 \gain{+6.5}
    & 84.1 \gain{+9.7}
    & 76.4 \gain{+9.5}
    & 75.9 \gain{+9.0} \\
    Progress-SQL-14B(Ours) & Progressive 
    & 77.1 \gain{+4.8}
    & 69.1 \gain{+3.5}
    & 84.1 \gain{+2.6}
    & 76.6 \gain{+2.0}
    & 76.4 \gain{+4.1} \\
    \bottomrule
  \end{tabular*}
  \caption{
  Robustness evaluation on Spider variants.
  All comparative 7B RL baselines are based on Qwen2.5-Coder-7B-Instruct.
  We report EX and TS if they are available.
  Numbers in red fonts denote absolute gains over the same-scale base model.
  $^\dagger$ indicates results reproduced by us following the data construction
described in the corresponding papers and without additional task-specific
supervised fine-tuning.
  }
  \label{tab:robustness_results}
\end{table*}

\subsection{Further Analysis}
\label{subsec:robustness_evaluation}

\noindent \textbf{Robustness Evaluation}. Table~\ref{tab:robustness_results} presents the robustness evaluation on Spider-Syn, Spider-Realistic, and Spider-DK, which include various knowledge perturbation. 
Our models consistently improves over the corresponding Qwen2.5-Coder base models across all three variants, which showcases the robustness of our method on different Text-to-SQL scenarios.
Moreover, our models achieve competitive results compared with existing post-training methods across the Spider variants, suggesting superior effect of post-training for Text-to-SQL.



\begin{figure*}[!t]
    \centering

    \begin{subfigure}[t]{0.48\textwidth}
        \centering
        \includegraphics[width=\linewidth]{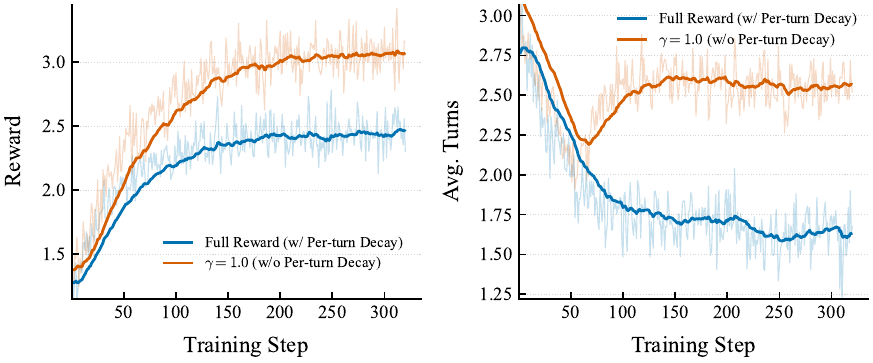}
        \caption{Effects of per-turn decay.}
        \label{fig:reward_hacking}
    \end{subfigure}
    \hfill
    \begin{subfigure}[t]{0.48\textwidth}
        \centering
        \includegraphics[width=\linewidth]{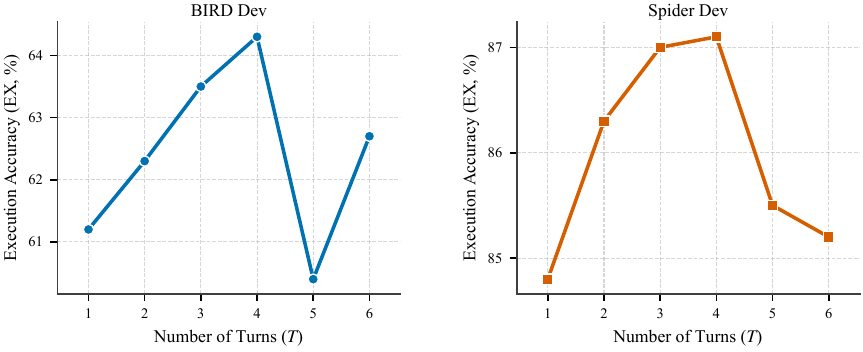}
        \caption{Effects of maximum interaction turns $T$.}
        \label{fig:ex_vs_turns}
    \end{subfigure}

    \caption{
    Effects of per-turn decay and interaction budget.
    (a) Removing per-turn decay leads to less stable multi-turn optimization and longer correction trajectories.
    (b) Performance improves up to $T=4$ and then decreases when more interaction turns are allowed.
    }
    \label{fig:decay_and_turn_ablation}
\end{figure*}

\begin{table}[!t]
  \centering
  \footnotesize
  \setlength{\tabcolsep}{3.4pt}
  \renewcommand{\arraystretch}{1.05}

  \begin{tabular*}{\columnwidth}{@{\extracolsep{\fill}}lccc@{}}
    \toprule
    \multirow{2}{*}{\textbf{Variant}} 
    & \textbf{BIRD} 
    & \multicolumn{2}{c}{\textbf{Spider}} \\
    \cmidrule(lr){2-2} \cmidrule(l){3-4}
    & \textbf{EX(\%)} & \textbf{EX(\%)} & \textbf{TS(\%)} \\
    \midrule
    \textbf{Full} 
    & \textbf{64.3} & \textbf{87.1} & \textbf{80.9} \\
    \midrule

    Single-turn RL
    & 61.2 \drop{-3.1} 
    & 84.8 \drop{-2.3} 
    & 76.9 \drop{-4.0} \\

    w/o $\mathcal{F}_{\mathrm{struct}}$
    & 61.6 \drop{-2.7} 
    & 85.5 \drop{-1.6} 
    & 78.4 \drop{-2.5} \\

    w/o $\mathcal{F}_{\mathrm{lex}}$
    & 62.1 \drop{-2.2} 
    & 85.9 \drop{-1.2} 
    & 78.8 \drop{-2.1} \\

    w/o $\mathcal{R}_{\mathrm{exec}}$
    & 63.1 \drop{-1.2} 
    & 86.5 \drop{-0.6} 
    & 78.1 \drop{-2.8} \\

    $\gamma=1$
    & 47.5 \drop{-16.8} 
    & 61.3 \drop{-25.8} 
    & 56.5 \drop{-24.4} \\

    \bottomrule
  \end{tabular*}

  \caption{Ablation study on BIRD Dev and Spider Dev under Vote@8 decoding. Numbers in blue fonts in parentheses denote absolute drops compared with the full reward.}
  \label{tab:ablation}
\end{table}

\noindent \textbf{Ablation Study}. We conduct an ablation study on BIRD Dev and Spider Dev. Table~\ref{tab:ablation} reports the results under Vote@8 decoding. 
As we can see, iteratively removing components from the comprehensive reward leads to performance drop on both benchmarks.
The Single-turn RL variant removes the multi-turn rollout and the
progressive alignment reward $\mathcal{R}_{\mathrm{align}}$, reducing our framework
to a standard one-shot RL setting where the model generates a single SQL and is
optimized using only final-state rewards. 
Among all the variants, removing per-turn decay causes the largest performance drop. This indicates that per-turn decay is critical for preventing the policy from over-relying on the late feedback.
Figure~\ref{fig:reward_hacking} further illustrates the training dynamics behind this degradation.
Without per-turn decay, the model is not penalized for solving the query in later turns and therefore tends to rely more heavily on ODT feedback to progressively repair its SQL. This increases the average number of interaction turns and allows the trajectory-level reward to improve, but weakens the model's first-attempt SQL generation ability. Consequently, when evaluated under the standard single-turn protocol without ODT feedback, the no-decay model performs substantially worse. More detailed analysis is provided in Appendix~\ref{app:per_turn_decay}.

\noindent \textbf{Impact of Maximum Interaction Turns}.
We study the maximum interaction budget $T$, which controls the number of refinement attempts in each rollout.
As shown in Figure~\ref{fig:ex_vs_turns}, performance improves as $T$ increases up to 4, suggesting that additional turns help correct SQL errors.
However, larger budgets lead to performance degradation, indicating that more refinement is not always beneficial.
Therefore, we set $T=4$ in all main experiments.
Detailed analysis is provided in Appendix~\ref{app:ex_turns}.

\section{Conclusion}

In this work, we proposed Progress-SQL, a multi-turn reinforcement learning framework with progressive rewards for Text-to-SQL.
Progress-SQL leverages ODT-based diagnostic feedback to guide iterative SQL refinement and optimizes a progressive reward that jointly encourages alignment improvement, early correctness, and recovery from invalid SQL.
Experiments on multiple Text-to-SQL benchmarks show consistent gains across primary and robustness evaluations, demonstrating the effectiveness of our framework for improving Text-to-SQL reasoning and refinement.

\section*{Limitations}
Despite the effectiveness of our framework, two limitations remain.
On the one hand, our ODT-based diagnostic feedback relies on the availability of gold SQL during training. Therefore, ODT is mainly designed as a training-time reinforcement learning signal rather than a test-time feedback module, since gold SQL is unavailable during standard inference. Nevertheless, our experiments show that such oracle-guided training feedback can be internalized by the policy model and improve its single-turn inference ability without using ODT at test time.
On the other hand, the ODT scorer depends on SQL parsing and structural abstraction. Although lexical alignment is introduced as a fallback signal for ill-formed SQL predictions, severely malformed queries or dialect-specific SQL constructs may still reduce the precision of structural diagnostics. Extending the framework to more robust SQL parsers, diverse SQL dialects, and real-world interactive database settings remains an important direction for future work.

\bibliography{custom}

\clearpage
\appendix

\section{Use of Large Language Models}
The research presented in this paper, including the core ideas, experimental design, and quantitative results, is the original work of the authors. A large language model was used as a writing assistant for tasks such as polishing prose, improving clarity, and correcting grammatical errors in the manuscript. All final content was reviewed and edited by the authors to ensure it accurately reflects our research and contributions.

\section{Implementation Details}
\label{app:implementation_details}

\noindent \textbf{Training Data Construction}
The RL training corpus is constructed from three sources: the \textbf{BIRD} training
set~\citep{DBLP:conf/nips/LiHQYLLWQGHZ0LC23}, the \textbf{Spider} training
set~\citep{DBLP:conf/emnlp/YuZYYWLMLYRZR18}, and \textbf{SynSQL-2.5M}~\citep{DBLP:journals/pvldb/LiWZHZJWZCSCL25}, a large-scale synthetic dataset covering diverse SQL complexity patterns.
We randomly sample a total of 20{,}000 training instances from the union of these sources for reinforcement learning.

\noindent \textbf{Training Setup and Infrastructure.}
We adopt Qwen2.5-Coder-7B/14B-Instruct as our base model for all reinforcement learning experiments. The training pipeline is developed and executed utilizing the \texttt{verl} framework across 8 NVIDIA H100 (80GB) GPUs. We employ Group Relative Policy Optimization (GRPO) as our core algorithm, which inherently bypasses the need for a separate value network (critic), thereby substantially reducing memory overhead and improving training efficiency. During the rollout phase, we sample $G=8$ outputs per input query with a temperature of $\tau = 1.0$ and a maximum sequence length of 8192 tokens. The maximum number of interaction turns with the database engine is set to $T=4$. The comprehensive hyperparameters for both the multi-turn rollout and policy optimization phases are summarized in Table~\ref{tab:hyperparameters}.

\noindent \textbf{Evaluation Setup.}
We follow the evaluation scripts and protocol released by OmniSQL~\citep{DBLP:journals/pvldb/LiWZHZJWZCSCL25}\footnote{\url{https://github.com/RUCKBReasoning/OmniSQL}} to ensure comparability with recent Text-to-SQL baselines.
At evaluation time, ODT-based diagnostic feedback is not used, and all models are evaluated under standard single-turn inference.

\noindent \textbf{Reward Hyperparameters.}
The scalar weights for our multi-turn reward formulation (defined in Section~\ref{subsec:reward_formulation}) are systematically detailed in Table~\ref{tab:hyperparameters}. 
\begin{table}[htbp]
\centering
\small
\setlength{\tabcolsep}{5pt}
\renewcommand{\arraystretch}{1.02}
\begin{tabularx}{\linewidth}{@{}X>{\centering\arraybackslash}p{0.22\linewidth}@{}}
\toprule
\textbf{Hyperparameter} & \textbf{Value} \\
\midrule
\multicolumn{2}{@{}l}{\textit{RL Training}} \\
Optimizer                          & AdamW \\
Learning Rate                      & $1\times10^{-6}$ \\
Train Batch Size                   & 64 \\
Sampled Outputs $G$                & 8 \\
Sampling Temperature $\tau$        & 1.0 \\
Max Response Length                & 8192 \\
Max Prompt Length                  & 6000 \\
Clip Ratio $\varepsilon$           & 0.2 \\
Max Interaction Turns $T$          & 4 \\
KL Penalty                         & None \\
\midrule
\multicolumn{2}{@{}l}{\textit{Reward Weights}} \\
Formatting Reward $\omega_{\mathrm{fmt}}$              & 0.5 \\
Base Accuracy Reward $\omega_{\mathrm{acc}}$           & 2.0 \\
Alignment Improvement $\omega_{\mathrm{align}}^{+}$    & 1.0 \\
Alignment Penalty $\omega_{\mathrm{align}}^{-}$        & -0.25 \\
Executable Keeping $\omega_{\mathrm{keep}}$            & 0.5 \\
Execution Recovery $\omega_{\mathrm{rec}}$             & 0.25 \\
Execution Deterioration $\omega_{\mathrm{det}}$        & -0.25 \\
Per-turn Decay Factor $\gamma$                         & 0.5 \\
\bottomrule
\end{tabularx}
\caption{Hyperparameters for RL training and progressive reward formulation.}
\label{tab:hyperparameters}
\end{table}

\section{Multi-turn Rollout with ODT Feedback}
\label{app:rollout_algorithm}

Algorithm~\ref{alg:multi_turn_rollout} summarizes the training-time rollout procedure.
At each turn, the policy generates a SQL prediction and receives ODT-based diagnostic feedback if the prediction is not yet correct.
Here, $T$ denotes the maximum turn budget, $k^*$ denotes the first successful turn if execution correctness is reached, and $K$ denotes the actual stopping turn.
If the rollout succeeds early, then $K=k^*$; otherwise, $K=T$.

\begin{algorithm*}[t]
\caption{Multi-turn SQL Refinement with ODT Feedback}
\label{alg:multi_turn_rollout}
\small
\begin{algorithmic}[1]
\REQUIRE Question $x$, database $D$, gold SQL $y^*$, policy model $\pi_\theta$, maximum turns $T$
\ENSURE Trajectory $Y=\{y^{(1)},\ldots,y^{(K)}\}$ and final SQL $y^{(K)}$, where $K \leq T$

\STATE Initialize prompt $p^{(1)} \leftarrow x$, trajectory $Y \leftarrow \emptyset$
\STATE Initialize first successful turn $k^* \leftarrow \varnothing$ and stopping turn $K \leftarrow T$

\FOR{$t = 1$ to $T$}
    \STATE Generate SQL prediction $y^{(t)} \sim \pi_\theta(\cdot \mid p^{(t)})$ and append it to $Y$
    \STATE Execute $y^{(t)}$ on database $D$ to obtain execution status $\mathcal{E}(y^{(t)})$

    \IF{$y^{(t)}$ matches the execution result of gold SQL $y^*$}
        \STATE Set $k^* \leftarrow t$ and $K \leftarrow t$; \textbf{break}
    \ENDIF

    \IF{$t < T$}
        \STATE Compare $y^{(t)}$ with $y^*$ using the ODT diagnostic module
        \STATE Obtain clause-level diagnostic tags and verbalize them into feedback $f^{(t)}$
        \STATE Update prompt $p^{(t+1)} \leftarrow \mathrm{Append}(p^{(t)}, y^{(t)}, f^{(t)})$
    \ENDIF
\ENDFOR

\STATE Compute alignment improvement $\Delta = \mathcal{F}(y^{(K)}, y^*) - \mathcal{F}(y^{(1)}, y^*)$
\STATE Compute total reward using alignment, decayed accuracy, execution status, and formatting terms
\RETURN Trajectory $Y$ and final SQL $y^{(K)}$
\end{algorithmic}
\end{algorithm*}

\section{ODT Construction and Scoring}
\label{app:odt_scoring}

\noindent \textbf{Implementation Overview.}
We implement the Oracle-guided Diagnostic Tree (ODT) module as a deterministic structural feedback component.
In our implementation, SQL queries are parsed with JSQLParser 4.6 and converted into recursive structural profile trees.
Given a predicted SQL $y$ and the gold SQL $y^*$, ODT extracts clause-level structural profiles, computes a structural similarity score, and generates diagnostic tags for next-turn feedback.
The diagnostic feedback is used only during multi-turn RL training.
At evaluation time, the model performs standard single-turn inference without ODT feedback.

\noindent \textbf{SQL Parsing and Failure Handling.}
For each generated SQL, ODT first parses the query into an Abstract Syntax Tree (AST).
Before parsing, we apply lightweight normalization, including removing redundant quotation marks, standardizing whitespace, lower-casing identifiers, and stripping trailing semicolons.
If the predicted SQL cannot be parsed, ODT does not produce fine-grained structural tags.
Instead, it returns a generic feedback message asking the model to first fix syntax and identifier errors.
In this case, the structural similarity is set to zero, and the reward relies on lexical alignment and execution-status signals.

\noindent \textbf{ODT Construction.}
Each parsed SQL is converted into a recursive ODT profile tree.
Each node corresponds to a structural unit, including \texttt{ROOT}, \texttt{SELECT}, \texttt{CTE}, \texttt{SET\_OP}, or \texttt{SUBQUERY}.
A \texttt{SELECT} node contains a local structural profile.
A \texttt{CTE} node represents a common table expression, a \texttt{SET\_OP} node represents set operations such as \texttt{UNION}, \texttt{INTERSECT}, or \texttt{EXCEPT}, and a \texttt{SUBQUERY} node represents a nested query.
Subqueries appearing in \texttt{FROM}, \texttt{WHERE}, \texttt{HAVING}, or SELECT expressions are recursively detected and attached as child nodes.
This representation allows ODT to compare both flat clause-level structures and nested SQL compositions.

\noindent \textbf{Local Structural Profile.}
For each \texttt{SELECT} node, ODT extracts a local structural profile containing selected columns, involved tables, join signatures, filtering predicates, grouping columns, having predicates, ordering signatures, DISTINCT usage, aggregation usage, and structural counts.
Concretely, the profile contains:
\[
p = (\mathbf{c}, \mathbf{b}, \mathcal{T}, \mathcal{S}, \mathcal{W}, 
\mathcal{J}, \mathcal{G}, \mathcal{H}, \mathcal{O}),
\]
where $\mathbf{c}$ denotes count features such as JOIN count and SELECT-item count, and $\mathbf{b}$ denotes Boolean indicators for WHERE, GROUP BY, HAVING, ORDER BY, LIMIT, aggregation, and DISTINCT.
The set $\mathcal{T}$ contains normalized table names, $\mathcal{S}$ contains SELECT projection signatures, $\mathcal{W}$ contains normalized WHERE predicates, $\mathcal{J}$ contains JOIN-condition signatures, $\mathcal{G}$ contains GROUP BY column signatures, $\mathcal{H}$ contains HAVING predicates, and $\mathcal{O}$ contains ORDER BY signatures.

\noindent \textbf{Feature Normalization.}
To reduce sensitivity to surface-level SQL variation, ODT normalizes structural features before comparison.
Table names and column references are lower-cased, and quotation marks are removed.
String and numeric literals are replaced with placeholders.
Common predicate operators, including \texttt{=}, \texttt{<>}, \texttt{!=}, \texttt{>=}, \texttt{<=}, \texttt{>}, \texttt{<}, \texttt{LIKE}, \texttt{IN}, \texttt{BETWEEN}, \texttt{IS NULL}, and \texttt{IS NOT NULL}, are mapped into canonical predicate signatures.
For equality predicates where both sides are identifiers, the operands are sorted lexicographically, so \texttt{a=b} and \texttt{b=a} are treated as equivalent.
Conjunctive predicates are split by \texttt{AND} and stored as sets of normalized atomic predicates, making the order of independent conditions irrelevant.
Aggregation aliases are ignored during structural comparison; for example, \texttt{AVG(price) AS avg\_price} and \texttt{AVG(price) AS average\_price} are treated as equivalent if the aggregation function and argument match.

\noindent \textbf{Structural Scoring.}
Given the predicted ODT and the gold ODT, we recursively compute node-level similarity.
For each \texttt{SELECT} node, the local similarity is computed as a weighted average over feature-level similarities, with the feature weights listed in Table~\ref{tab:odt_scoring_weights}:
\[
s_{\mathrm{local}}(p,g)
=
\frac{
\sum_{f \in \mathcal{F}} w_f \cdot \mathrm{sim}_f(p_f,g_f)
}{
\sum_{f \in \mathcal{F}} w_f
}.
\]
For set-valued features, we use Jaccard similarity:
\[
\mathrm{Jaccard}(A,B)
=
\frac{|A \cap B|}{|A \cup B|}.
\]
If both sets are empty, the similarity is set to $1.0$; if only one side is empty, it is set to $0.0$.
For count-valued features, including JOIN count and SELECT-item count, we use:
\[
\mathrm{NumSim}(a,b)
=
1.0 -
\min\left(
1.0,
\frac{|a-b|}{\max(1,b)}
\right).
\]
For Boolean features such as DISTINCT usage, the similarity is $1.0$ if the two sides match and $0.0$ otherwise.

\begin{table}[htbp]
\centering
\small
\setlength{\tabcolsep}{5pt}
\renewcommand{\arraystretch}{1.02}
\begin{tabularx}{\linewidth}{@{}X>{\centering\arraybackslash}p{0.22\linewidth}@{}}
\toprule
\textbf{Feature} & \textbf{Weight} \\
\midrule
JOIN conditions      & 0.22 \\
WHERE predicates     & 0.20 \\
SELECT projections   & 0.16 \\
FROM/JOIN tables     & 0.12 \\
GROUP BY columns     & 0.10 \\
Structural counts    & 0.10 \\
ORDER BY clauses     & 0.06 \\
DISTINCT usage       & 0.04 \\
\bottomrule
\end{tabularx}
\caption{Feature weights used in ODT structural scoring.}
\label{tab:odt_scoring_weights}
\end{table}

\noindent \textbf{Recursive Child Matching.}
The final score of an ODT node combines local similarity and child-subtree similarity:
\[
s_{\mathrm{node}} =
\alpha \cdot s_{\mathrm{local}}
+
(1-\alpha) \cdot s_{\mathrm{child}}.
\]
We set $\alpha=0.70$ for \texttt{SELECT} nodes.
For non-\texttt{SELECT} nodes, which do not contain local profiles, the score is determined by child matching.
To compute $s_{\mathrm{child}}$, ODT matches predicted and gold children with compatible node types.
For CTE nodes, the CTE names are matched case-insensitively.
For each gold child, the scorer selects the highest-scoring unmatched predicted child.
The child score is normalized by the maximum number of children on the two sides:
\[
s_{\mathrm{child}}
=
\frac{
\sum_{(u,v) \in \mathcal{M}} s_{\mathrm{node}}(u,v)
}{
\max(|\mathcal{C}_{\mathrm{pred}}|, |\mathcal{C}_{\mathrm{gold}}|)
},
\]
where $\mathcal{M}$ is the set of matched child pairs.
Unmatched gold children indicate missing nested structures, while unmatched predicted children indicate redundant subqueries or CTEs.
The final structural similarity is the root score:
\[
\mathcal{F}_{\mathrm{struct}}(y,y^*) = s_{\mathrm{root}}.
\]

\noindent \textbf{Diagnostic Tag Generation.}
In addition to the continuous structural similarity score, ODT produces discrete diagnostic tags for feedback construction.
For each diagnostic dimension, we compare the corresponding feature similarity against a predefined threshold, as listed in Table~\ref{tab:odt_thresholds}.
If the similarity falls below the threshold, the corresponding error tag is emitted.
Table~\ref{tab:odt_tags} summarizes the diagnostic tags and their meanings.
These tags are then verbalized into concise natural-language feedback and appended to the next-turn prompt.
The raw structural similarity score and feature-level similarities are not exposed to the model; they are used only for reward computation.

\begin{table}[htbp]
\centering
\small
\setlength{\tabcolsep}{5pt}
\renewcommand{\arraystretch}{1.02}
\begin{tabularx}{\linewidth}{@{}X>{\centering\arraybackslash}p{0.26\linewidth}@{}}
\toprule
\textbf{Diagnostic Dimension} & \textbf{Threshold} \\
\midrule
JOIN conditions      & 0.60 \\
WHERE predicates     & 0.60 \\
SELECT projections   & 0.60 \\
GROUP BY columns     & 0.65 \\
ORDER BY clauses     & 0.65 \\
FROM/JOIN tables     & 0.70 \\
\bottomrule
\end{tabularx}
\caption{Mismatch thresholds for generating ODT diagnostic tags.}
\label{tab:odt_thresholds}
\end{table}

\begin{table*}[htbp]
\centering
\small
\setlength{\tabcolsep}{4pt}
\renewcommand{\arraystretch}{1.08}
\begin{tabularx}{\textwidth}{@{}lXX@{}}
\toprule
\textbf{Tag} & \textbf{Trigger} & \textbf{Meaning} \\
\midrule
\texttt{JOIN\_MISSING}
& JOIN count differs from the reference.
& Required JOIN structures are missing. \\

\texttt{JOIN\_KEY\_MISMATCH}
& JOIN-condition similarity is below threshold.
& JOIN keys or ON conditions are inconsistent. \\

\texttt{FROM\_OR\_JOIN\_TABLE\_MISMATCH}
& Table-set similarity is below threshold.
& Missing, redundant, or incorrect tables are used. \\

\texttt{WHERE\_ERROR}
& WHERE-predicate similarity is below threshold.
& Filtering predicates differ from the reference. \\

\texttt{GROUP\_BY\_MISSING}
& Gold SQL contains GROUP BY columns but the predicted SQL has no GROUP BY clause.
& Required grouping structure is missing. \\

\texttt{GROUP\_BY\_ERROR}
& GROUP-BY similarity is below threshold.
& Grouping columns or granularity are inconsistent. \\

\texttt{ORDER\_BY\_MISMATCH}
& ORDER-BY similarity is below threshold.
& Ordering column, direction, or limit structure is inconsistent. \\

\texttt{SELECT\_ERROR}
& SELECT-projection similarity is below threshold.
& Selected columns or aggregation expressions differ. \\

\texttt{AGGREGATE\_ERROR}
& Aggregation usage differs.
& Aggregation functions may be missing or misused. \\

\texttt{DISTINCT\_MISMATCH}
& DISTINCT usage differs.
& Duplicate-removal behavior is inconsistent. \\

\texttt{SUBQUERY\_MISSING}
& A gold subquery child has no compatible match.
& An expected nested query branch is missing. \\

\texttt{EXTRA\_SUBQUERY\_OR\_CTE}
& Predicted tree contains unmatched extra children.
& Redundant nested or CTE branches are introduced. \\
\bottomrule
\end{tabularx}
\caption{Diagnostic tags emitted by the ODT module.}
\label{tab:odt_tags}
\end{table*}

\noindent \textbf{Usage During RL Rollouts.}
For each unsuccessful non-final rollout turn, ODT compares the current prediction $y^{(t)}$ with the gold SQL $y^*$ and returns verbalized structural feedback $f^{(t)}$.
The next-turn input is constructed as:
\[
x^{(t+1)}
=
(q,S,y^{(1)},f^{(1)},\ldots,y^{(t)},f^{(t)}).
\]
ODT is a fixed, non-differentiable component of the training environment.
It does not participate in back-propagation, and gradients are computed only through the policy model over generated tokens.

\section{Additional Analysis}
\label{app:additional_analysis}

\subsection{Training Dynamics}
\label{app:training_trends}

\begin{figure}[h]
    \centering
    \includegraphics[width=\columnwidth]{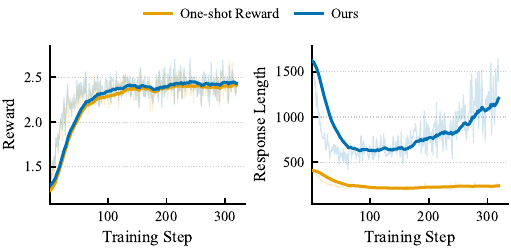}
    \caption{Training dynamics of reward and response length.
The single-turn one-shot reward baseline assigns rewards based on execution
correctness, executability, and format compliance, while our method further
incorporates multi-turn refinement, ODT-based structural feedback, and
trajectory-level progress. Both methods show stable reward improvement, but our
progressive reward leads to different response-length dynamics that are more
consistent with multi-turn SQL refinement.}
    \label{fig:reward_length}
\end{figure}

To better understand the impact of our progressive reward formulation on optimization dynamics, we analyze the training curves of reward and response length.
Figure~\ref{fig:reward_length} compares our full multi-turn progressive reward with a single-turn one-shot reward baseline.
The single-turn baseline generates one SQL prediction per rollout and assigns a score of $2.0$ if the generated SQL matches the gold execution result, $0.5$ if the SQL is executable, and $0.5$ if the output follows the required format.
Unlike our method, this baseline does not use multi-turn refinement, ODT-based structural feedback, or trajectory-level improvement.

\noindent \textbf{Reward Convergence.}
As shown in the left panel of Figure~\ref{fig:reward_length}, both the single-turn
one-shot baseline and our progressive reward improve steadily during training
and reach comparable reward levels. This indicates that our multi-turn objective
does not make optimization harder despite introducing ODT-based feedback and
trajectory-level reward components. Compared with the single-turn baseline, our
reward provides additional structural and lexical alignment signals that guide
the policy toward refinement-oriented behavior rather than only endpoint
correctness.

\noindent \textbf{Different Response-Length Dynamics.}
The right panel of Figure~\ref{fig:reward_length} shows that the two methods lead
to clearly different response-length behaviors. The single-turn one-shot baseline
quickly converges to shorter responses, since it only rewards final correctness,
executability, and format compliance. In contrast, our method maintains longer
responses during training, which is consistent with the multi-turn refinement
setting where the model needs to reason about previous SQL attempts and
diagnostic feedback. Although longer responses are not directly rewarded, this
suggests that the policy learns to utilize the additional context and feedback
rather than collapsing to a minimal endpoint-oriented generation strategy.

Overall, these trends show that the proposed progressive reward achieves stable
optimization while inducing training dynamics better aligned with multi-turn SQL
correction.

\subsection{Effect of Per-turn Decay}
\label{app:per_turn_decay}
Figure~\ref{fig:reward_hacking} analyzes the effect of removing the per-turn
decay term by setting $\gamma=1.0$. Although the accumulated reward continues to
increase, the average number of interaction turns first decreases and then
rebounds to a high level. This suggests that removing per-turn decay makes
multi-turn RL optimization unstable: the policy is no longer encouraged to
produce correct SQL early, and instead tends to rely on longer correction
trajectories with more ODT feedback.

This instability also explains the large performance drop in
Table~\ref{tab:ablation}. Without per-turn decay, late corrections receive the
same accuracy reward as early corrections, which weakens the pressure to improve
the first-attempt SQL. As a result, the learned policy becomes poorly aligned
with the standard single-turn inference setting used at evaluation time.
The per-turn decay term therefore acts as an important regularizer for
multi-turn RL training, encouraging earlier successful executions and stabilizing
the transfer from multi-turn training to single-turn inference.

\subsection{Maximum Interaction Turns}
\label{app:ex_turns}
Figure~\ref{fig:ex_vs_turns} shows the effect of varying the maximum interaction budget $T$.
Increasing $T$ from 1 to 4 consistently improves execution accuracy on both BIRD Dev and Spider Dev, suggesting that additional turns provide useful opportunities for structural correction.
However, further increasing $T$ causes performance degradation.
We attribute this to context accumulation and over-correction: longer trajectories introduce more historical SQL attempts and feedback messages, which may distract the model or make it deviate from the original question.
Based on these observations, we set $T=4$ in all main experiments.

\begin{table*}[t]
\centering
\footnotesize
\setlength{\tabcolsep}{4pt}
\renewcommand{\arraystretch}{1.08}
\newcommand{\errchg}[2]{%
  #1~{\scriptsize#2}%
}
\begin{tabular*}{0.96\textwidth}{@{}l@{\extracolsep{\fill}}ccccc@{}}
\toprule
\textbf{Model} 
& \textbf{SELECT Err.} 
& \textbf{JOIN Err.} 
& \textbf{WHERE Err.} 
& \textbf{GROUP Err.} 
& \textbf{Avg. Err.} \\
\midrule
Qwen2.5-Coder-7B 
& 47.3 & 38.1 & 33.6 & 19.3 & 34.6 \\
Progress-SQL-7B 
& \errchg{42.0}{$(\downarrow5.3)$}
& \errchg{29.9}{$(\downarrow8.2)$}
& \errchg{30.7}{$(\downarrow2.9)$}
& \errchg{13.4}{$(\downarrow5.9)$}
& \errchg{29.0}{$(\downarrow5.6)$} \\
\midrule
Qwen2.5-Coder-14B 
& 68.6 & 17.5 & 39.2 & 23.3 & 37.1 \\
Progress-SQL-14B 
& \errchg{37.8}{$(\downarrow30.8)$}
& \errchg{24.1}{$(\uparrow6.6)$}
& \errchg{31.0}{$(\downarrow8.2)$}
& \errchg{11.5}{$(\downarrow11.8)$}
& \errchg{26.1}{$(\downarrow11.0)$} \\
\bottomrule
\end{tabular*}
\caption{ODT-based structural mismatch rates for single-turn greedy inference on Spider Dev.
The ODT diagnostic module is used only as an offline evaluator and is not exposed to the model during inference.
These rates measure clause-level structural mismatches rather than execution failures, and thus are not directly complementary to EX.
Arrows indicate absolute changes compared with the corresponding same-scale base model; lower values are better.
Avg. Err. averages SELECT, JOIN, WHERE, and GROUP mismatch rates.}
\label{tab:single_turn_odt_error_rate}
\end{table*}

\subsection{Single-turn Inference Behavior after Multi-turn RL}
\label{app:single_turn_inference}

During training, our framework exposes the policy to multi-turn ODT-based diagnostic feedback, while during evaluation we deliberately use standard single-turn inference without ODT feedback or iterative refinement. This setting ensures fair comparison with existing Text-to-SQL baselines, but also raises an important question: how can a model trained in a multi-turn debugging environment benefit at single-turn test time?

We hypothesize that multi-turn RL does not merely teach the model to react to external feedback, but also helps the policy internalize structural debugging patterns. In other words, after optimization with progressive rewards, the model may learn to anticipate common SQL construction errors before producing the final query. Such internalization should be reflected in fewer structural errors in the first generated SQL, even when no ODT feedback is provided at inference time.

To verify this hypothesis, we apply the ODT diagnostic module only as an offline evaluator for single-turn greedy predictions on Spider Dev. Specifically, after each model generates its SQL, we compare the predicted SQL with the gold SQL using ODT and map fine-grained diagnostic tags into four coarse structural categories: SELECT, JOIN, WHERE, and GROUP errors. Importantly, this analysis is performed after generation; the diagnostic tags are never exposed to the model during evaluation.

As shown in Table~\ref{tab:single_turn_odt_error_rate}, RL-trained models consistently reduce the average structural error rate under single-turn inference on Spider Dev. For the 7B model, the average error rate decreases from 34.6\% to 29.0\%, with especially clear reductions in JOIN and GROUP errors. For the 14B model, the average error rate decreases from 37.1\% to 26.1\%, mainly due to large reductions in SELECT, WHERE, and GROUP errors. Although the JOIN error rate of the 14B model increases, the overall structural error rate still drops substantially. These results suggest that the benefit of multi-turn ODT-based training is not limited to interactive correction: the learned policy also improves first-attempt SQL construction without relying on test-time oracle feedback.


\subsection{Decoding Strategy Analysis by Difficulty}
\label{app:decoding_difficulty}

To provide a fine-grained view of the inference behavior of our RL-trained models, we report their performance across SQL query difficulty levels under two decoding strategies.
Specifically, we evaluate Progress-SQL-7B and Progress-SQL-14B on BIRD Dev, Spider Dev, Spider Test, and three robustness variants: Spider-DK, Spider-Realistic, and Spider-Syn.
For each benchmark, we include both Greedy Decoding and Majority Voting (Vote@8).

\noindent \textbf{Overall Observation.}
The results show that decoding strategy can affect performance differently across benchmarks and difficulty subsets.
Vote@8 often changes the aggregate score compared with greedy decoding, but the effect is not uniform across all settings.
In some hard or extra-hard subsets, majority voting may select a suboptimal candidate when generated SQLs differ in subtle structural details.
Therefore, this analysis is intended to provide a more detailed view of decoding behavior rather than to claim that one decoding strategy consistently dominates the other.

\noindent \textbf{BIRD Dev.}
Table~\ref{tab:app_bird_decoding} reports the BIRD Dev results.
We report both greedy decoding and Vote@8 results for Progress-SQL-7B and Progress-SQL-14B, showing how performance varies across simple, moderate, and challenging examples.

\begin{table}[htbp]
  \centering
  \small
  \setlength{\tabcolsep}{4.2pt}
  \renewcommand{\arraystretch}{1.08}

  \resizebox{\linewidth}{!}{
  \begin{tabular}{@{}llcccc@{}}
    \toprule
    \multirow{2}{*}{\textbf{Model}} 
    & \multirow{2}{*}{\textbf{Decoding}}
    & \multicolumn{4}{c}{\textbf{BIRD Dev (EX\%)}} \\
    \cmidrule(l){3-6}
    & & \textbf{Simple} & \textbf{Moderate} & \textbf{Chall.} & \textbf{All} \\
    \midrule

    \multirow{2}{*}{Progress-SQL-7B}
    & Greedy & 68.6 & 54.5 & 49.7 & 62.6 \\
    & Vote@8 & 71.2 & 55.2 & 49.0 & 64.3 \\

    \midrule

    \multirow{2}{*}{Progress-SQL-14B}
    & Greedy & 72.2 & 58.0 & 53.8 & 66.2 \\
    & Vote@8 & 73.8 & 58.4 & 55.9 & 67.5 \\

    \bottomrule
  \end{tabular}
  }

  \caption{Execution accuracy breakdown by difficulty on BIRD Dev for our RL-trained models under greedy decoding and Vote@8.}
  \label{tab:app_bird_decoding}
\end{table}


\noindent \textbf{Spider Dev and Spider Test.}
Table~\ref{tab:app_spider_decoding} reports the results on Spider Dev and Spider Test.
We include both greedy decoding and Vote@8 results to show how decoding choices affect different difficulty subsets.
The aggregate scores are generally close between the two decoding strategies, while the difficulty-level breakdown reveals non-uniform changes across easy, medium, hard, and extra-hard examples.
In particular, Vote@8 does not consistently improve hard or extra-hard subsets, suggesting that majority voting may select a suboptimal candidate when generated SQLs differ in subtle structural details.

\begin{table*}[!t]
\centering
\small
\renewcommand{\arraystretch}{1.08}
\setlength{\tabcolsep}{3.2pt}
\resizebox{\textwidth}{!}{
\begin{tabular}{ll ccccc ccccc ccccc}
\toprule
\multirow{3}{*}{\textbf{Model}} 
& \multirow{3}{*}{\textbf{Decoding}}
& \multicolumn{10}{c}{\textbf{Spider Dev}}
& \multicolumn{5}{c}{\textbf{Spider Test}} \\
\cmidrule(lr){3-12} \cmidrule(lr){13-17}
& & \multicolumn{5}{c}{\textbf{EX(\%)}}
  & \multicolumn{5}{c}{\textbf{TS(\%)}}
  & \multicolumn{5}{c}{\textbf{EX(\%)}} \\
\cmidrule(lr){3-7} \cmidrule(lr){8-12} \cmidrule(lr){13-17}
& & \textbf{Easy} & \textbf{Med.} & \textbf{Hard} & \textbf{Extra} & \textbf{All}
  & \textbf{Easy} & \textbf{Med.} & \textbf{Hard} & \textbf{Extra} & \textbf{All}
  & \textbf{Easy} & \textbf{Med.} & \textbf{Hard} & \textbf{Extra} & \textbf{All} \\
\midrule

\multirow{2}{*}{Progress-SQL-7B}
  & Greedy
    & 96.0 & 90.4 & 82.2 & 66.9 & 86.6
    & 95.2 & 85.9 & 71.8 & 50.0 & 80.0
    & 93.8 & 90.2 & 82.9 & 74.2 & 86.8 \\
  & Vote@8
    & 96.8 & 90.8 & 81.6 & 68.7 & 87.1
    & 95.6 & 87.2 & 71.8 & 51.2 & 80.9
    & 93.0 & 91.0 & 84.2 & 78.2 & 87.8 \\

\midrule

\multirow{2}{*}{Progress-SQL-14B}
  & Greedy
    & 95.6 & 90.6 & 82.2 & 68.7 & 86.8
    & 93.5 & 86.8 & 74.7 & 46.4 & 79.9
    & 94.7 & 90.5 & 82.9 & 77.3 & 87.6 \\
  & Vote@8
    & 96.0 & 92.4 & 81.0 & 67.5 & 87.3
    & 94.8 & 86.1 & 74.7 & 49.4 & 80.4
    & 94.9 & 91.2 & 83.6 & 77.3 & 88.1 \\

\bottomrule
\end{tabular}
}
\caption{Execution accuracy (EX\%) and test-suite accuracy (TS\%) breakdown by difficulty on Spider Dev and Spider Test for our RL-trained models under greedy decoding and Vote@8. Spider Test reports EX only.}
\label{tab:app_spider_decoding}
\end{table*}


\noindent \textbf{Robustness Variants.}
Tables~\ref{tab:app_spider_robustness_decoding} and~\ref{tab:app_spider_robustness_ts_decoding} report difficulty-level results on Spider-DK, Spider-Realistic, and Spider-Syn.
These results further show that the effect of decoding strategy varies across robustness settings and difficulty levels.
Therefore, we present both greedy and Vote@8 results to provide a more complete view of inference behavior under different perturbation scenarios.

\begin{table*}[t]
\centering
\small
\renewcommand{\arraystretch}{1.08}
\setlength{\tabcolsep}{3.2pt}
\resizebox{\textwidth}{!}{
\begin{tabular}{ll ccccc ccccc ccccc}
\toprule
\multirow{2}{*}{\textbf{Model}} 
& \multirow{2}{*}{\textbf{Decoding}}
& \multicolumn{5}{c}{\textbf{Spider-DK (EX\%)}}
& \multicolumn{5}{c}{\textbf{Spider-Realistic (EX\%)}}
& \multicolumn{5}{c}{\textbf{Spider-Syn (EX\%)}} \\
\cmidrule(lr){3-7} \cmidrule(lr){8-12} \cmidrule(lr){13-17}
& & \textbf{Easy} & \textbf{Med.} & \textbf{Hard} & \textbf{Extra} & \textbf{All}
  & \textbf{Easy} & \textbf{Med.} & \textbf{Hard} & \textbf{Extra} & \textbf{All}
  & \textbf{Easy} & \textbf{Med.} & \textbf{Hard} & \textbf{Extra} & \textbf{All} \\
\midrule

\multirow{2}{*}{Progress-SQL-7B}
  & Greedy
    & 82.7 & 77.6 & 67.6 & 60.0 & 73.8
    & 89.9 & 91.6 & 80.8 & 61.9 & 83.5
    & 87.1 & 78.6 & 71.2 & 55.0 & 75.5 \\
  & Vote@8
    & 82.7 & 79.7 & 73.0 & 61.9 & 75.9
    & 91.7 & 91.1 & 83.8 & 60.8 & 84.1
    & 89.1 & 79.5 & 71.8 & 55.6 & 76.6 \\

\midrule

\multirow{2}{*}{Progress-SQL-14B}
  & Greedy
    & 90.0 & 80.5 & 58.1 & 67.6 & 76.8
    & 96.3 & 91.1 & 77.8 & 64.9 & 84.4
    & 87.9 & 82.0 & 70.6 & 53.3 & 76.8 \\
  & Vote@8
    & 87.3 & 82.5 & 62.2 & 61.0 & 76.4
    & 96.3 & 91.1 & 75.8 & 63.9 & 84.1
    & 88.3 & 81.6 & 70.6 & 55.6 & 77.1 \\

\bottomrule
\end{tabular}
}
\caption{Execution accuracy (EX\%) breakdown by difficulty on Spider-DK, Spider-Realistic, and Spider-Syn for our RL-trained models under greedy decoding and Vote@8.}
\label{tab:app_spider_robustness_decoding}
\end{table*}

\begin{table*}[!t]
\centering
\small
\setlength{\tabcolsep}{3.6pt}
\renewcommand{\arraystretch}{1.05}

\begin{tabular*}{\textwidth}{@{\extracolsep{\fill}}llccccc@{\qquad}ccccc@{}}
\toprule
\multirow{2}{*}{\textbf{Model}} 
& \multirow{2}{*}{\textbf{Dec.}}
& \multicolumn{5}{c}{\textbf{Spider-Realistic (TS\%)}}
& \multicolumn{5}{c}{\textbf{Spider-Syn (TS\%)}} \\
\cmidrule(lr){3-7} \cmidrule(l){8-12}
& & \textbf{Easy} & \textbf{Med.} & \textbf{Hard} & \textbf{Extra} & \textbf{All}
  & \textbf{Easy} & \textbf{Med.} & \textbf{Hard} & \textbf{Extra} & \textbf{All} \\
\midrule
\multirow{2}{*}{Progress-SQL-7B}
  & Greedy
  & 90.8 & 84.7 & 70.7 & 41.2 & 75.0
  & 82.3 & 74.1 & 62.7 & 33.7 & 67.5 \\
  & Vote@8
  & 92.7 & 85.7 & 73.7 & 41.2 & 76.4
  & 84.3 & 75.0 & 62.7 & 36.1 & 68.8 \\
\midrule
\multirow{2}{*}{Progress-SQL-14B}
  & Greedy
  & 94.5 & 85.2 & 65.7 & 46.4 & 76.0
  & 84.3 & 75.0 & 63.8 & 34.3 & 68.7 \\
  & Vote@8
  & 94.5 & 86.7 & 67.7 & 44.3 & 76.6
  & 83.5 & 75.5 & 64.4 & 36.1 & 69.1 \\
\bottomrule
\end{tabular*}

\caption{Test-suite accuracy (TS\%) breakdown by difficulty on Spider-Realistic and Spider-Syn for our RL-trained models under greedy decoding and Vote@8. Spider-DK is omitted because test-suite accuracy is unavailable.}
\label{tab:app_spider_robustness_ts_decoding}
\end{table*}

\subsection{Generalization Across RL Algorithms}
\label{app:rl_generalization}

To verify that our proposed multi-turn progressive reward $\mathcal{R}(Y)$ is algorithm-agnostic, we evaluate it under two distinct RL training paradigms: \textbf{GRPO} (Group Relative Policy Optimization)~\citep{shao2024deepseekmathpushinglimitsmathematical}, our primary algorithm, and \textbf{GSPO} (Group Sequence Policy Optimization)~\citep{zheng2025groupsequencepolicyoptimization}. For a controlled comparison, both algorithms are initialized from the same Qwen2.5-Coder-7B-Instruct base model without any task-specific supervised fine-tuning, and are optimized using identical multi-turn reward signals.

For a given input query $x$, both algorithms sample a group of $G$ responses $\{y_i\}_{i=1}^G$ from the old policy $\pi_{\theta_{\text{old}}}(\cdot|x)$, and compute the group-based advantage estimation $\hat{A}_i$ using our multi-turn reward $\mathcal{R}(y_i)$:
\begin{equation*}
    \hat{A}_i = \frac{\mathcal{R}(y_i) - \text{mean}(\{\mathcal{R}(y_i)\}_{i=1}^G)}{\text{std}(\{\mathcal{R}(y_i)\}_{i=1}^G)}
\end{equation*}

\noindent \textbf{GRPO Formulation.} GRPO optimizes the policy at the token level. It computes the advantage for each token $y_{i,t}$ as $\hat{A}_{i,t} = \hat{A}_i$, and applies clipping to the token-level importance ratio $w_{i,t}(\theta) = \frac{\pi_\theta(y_{i,t}|x, y_{i,<t})}{\pi_{\theta_{\text{old}}}(y_{i,t}|x, y_{i,<t})}$:
\begin{equation*}
\begin{aligned}
    \mathcal{J}_{\text{GRPO}}(\theta) &= \mathbb{E}_{x \sim \mathcal{D}, \{y_i\}_{i=1}^G \sim \pi_{\theta_{\text{old}}}(\cdot|x)} \Bigg[ \frac{1}{G} \sum_{i=1}^G \frac{1}{|y_i|} \\
    & \quad \sum_{t=1}^{|y_i|} \min \Big( w_{i,t}(\theta) \hat{A}_{i,t}, \\
    & \quad \quad \text{clip} \big( w_{i,t}(\theta), 1 - \varepsilon, 1 + \varepsilon \big) \hat{A}_{i,t} \Big) \Bigg]
\end{aligned}
\end{equation*}

\noindent\textbf{GSPO Formulation.} In contrast, GSPO applies clipping to entire responses instead of individual tokens to better match sequence-level rewards. To reduce variance and control numerical range, it defines a length-normalized sequence importance ratio $s_i(\theta)$:
\begin{equation*}
\begin{aligned}
    s_i(\theta) &= \left( \frac{\pi_\theta(y_i|x)}{\pi_{\theta_{\text{old}}}(y_i|x)} \right)^{\frac{1}{|y_i|}} \\
    &= \exp \left( \frac{1}{|y_i|} \sum_{t=1}^{|y_i|} \log \frac{\pi_\theta(y_{i,t}|x, y_{i,<t})}{\pi_{\theta_{\text{old}}}(y_{i,t}|x, y_{i,<t})} \right)
\end{aligned}
\end{equation*}
The sequence-level optimization objective is then formulated as:
\begin{equation*}
\begin{aligned}
    \mathcal{J}_{\text{GSPO}}(\theta) &= \mathbb{E}_{x \sim \mathcal{D}, \{y_i\}_{i=1}^G \sim \pi_{\theta_{\text{old}}}(\cdot|x)} \Bigg[ \frac{1}{G} \sum_{i=1}^G \\
    & \quad \min \Big( s_i(\theta) \hat{A}_i, \\
    & \quad\text{clip} \big( s_i(\theta), 1 - \varepsilon, 1 + \varepsilon \big) \hat{A}_i \Big) \Bigg]
\end{aligned}
\end{equation*}

As shown in Table~\ref{tab:rl_generalization}, both GRPO and GSPO achieve strong performance under the same multi-turn progressive reward formulation, suggesting that our reward design is not tied to a specific RL optimizer. GRPO performs slightly better on BIRD Dev and Spider Dev TS, while GSPO obtains a marginally higher Spider Dev EX under Vote@8. These results indicate that the proposed progressive reward can transfer across different policy optimization algorithms, with only minor variations in final performance.

\begin{table}[htbp]
  \centering
  \footnotesize
  \setlength{\tabcolsep}{3.5pt}
  \renewcommand{\arraystretch}{1.08}

  \begin{tabular*}{\linewidth}{@{\extracolsep{\fill}}llccc@{}}
    \toprule
    \multirow{2}{*}{\textbf{Algorithm}}
    & \multirow{2}{*}{\textbf{Decoding}}
    & \textbf{BIRD Dev}
    & \multicolumn{2}{c}{\textbf{Spider Dev}} \\
    \cmidrule(lr){3-3} \cmidrule(l){4-5}
    & & \textbf{EX} & \textbf{EX} & \textbf{TS} \\
    \midrule
    GSPO & Greedy & 61.6 & 85.3 & 78.3 \\
         & Vote@8 & 63.8 & 87.5 & 79.9 \\
    \midrule
    GRPO & Greedy & 62.6 & 86.6 & 80.0 \\
         & Vote@8 & 64.3 & 87.1 & 80.9 \\
    \bottomrule
  \end{tabular*}

  \caption{Generalization of our multi-turn progressive reward $\mathcal{R}(Y)$ across RL algorithms. Both GRPO and GSPO are initialized from Qwen2.5-Coder-7B-Instruct and trained with identical reward signals.}
  \label{tab:rl_generalization}
\end{table}

\subsection{Generalization to Qwen3-8B}
\label{app:qwen3}

To further validate the generalizability of our framework, we apply the same training pipeline to Qwen3-8B~\citep{yang2025qwen3} without architecture-specific modification.

\begin{table}[htbp]
  \centering
  \small
  \setlength{\tabcolsep}{4pt}
  \renewcommand{\arraystretch}{1.08}

  \resizebox{\linewidth}{!}{
  \begin{tabular}{@{}llccc@{}}
    \toprule
    \multirow{2}{*}{\textbf{Model}} 
    & \multirow{2}{*}{\textbf{Decoding}}
    & \textbf{BIRD Dev}
    & \multicolumn{2}{c}{\textbf{Spider Dev}} \\
    \cmidrule(lr){3-3} \cmidrule(l){4-5}
    & & \textbf{EX(\%)} & \textbf{EX(\%)} & \textbf{TS(\%)} \\
    \midrule
    Qwen3-8B 
      & Greedy & 59.5 & 82.2 & 70.1 \\
      & Vote@8 & 62.3 & 84.6 & 72.0 \\
    \midrule
    Progress-SQL-8B
      & Greedy & 64.0\gain{+4.5} & 84.5\gain{+2.3} & 74.1\gain{+4.0} \\
      & Vote@8 & 65.8\gain{+3.5} & 85.9\gain{+1.3} & 74.9\gain{+2.9} \\
    \bottomrule
  \end{tabular}
  }

  \caption{Generalization results on Qwen3-8B. Subscript values denote absolute gains over the corresponding Qwen3-8B baseline.}
  \label{tab:qwen3}
\end{table}

As shown in Table~\ref{tab:qwen3}, our method consistently improves Qwen3-8B on both BIRD Dev and Spider Dev under greedy decoding and Vote@8. These results demonstrate that the proposed reward design is not tied to the Qwen2.5-Coder architecture. Instead, it generalizes effectively to Qwen3-8B, suggesting that our structural feedback and progressive reward formulation can benefit stronger base models as well. The consistent gains across both EX and TS suggest that our training objective improves not only final execution correctness, but also the robustness of the generated SQL programs.



\section{Prompt Templates}
\label{app:prompt-templates}

The following templates are used across all training and rollout phases. 
Placeholders are denoted in monospace format: \texttt{\{db\_engine\}} is filled with \texttt{SQLite}; \texttt{\{db\_details\}} contains the DDL schema of the target database; \texttt{\{question\}} is the natural-language query.

\begin{promptbox}{System Prompt}
\textbf{Task Overview:}\\
You are a data science expert. The user message contains the database schema and the natural language question. Your task is to understand the schema and generate a valid SQL query to answer the question.

\vspace{0.5em}
\noindent\textbf{Database Engine:}\\
\texttt{\{db\_engine\}}

\vspace{0.5em}
\noindent\textbf{Instructions:}
\begin{itemize}[leftmargin=1.5em, topsep=2pt, itemsep=0pt]
    \item \textbf{Dialect Strictness:} Ensure the generated SQL strictly conforms to the specific syntax and functions of the provided \texttt{\{db\_engine\}}.
    \item \textbf{Precision:} Only output the exact information asked in the question. Do not include extra columns in the \texttt{SELECT} clause.
    \item \textbf{Ambiguity Prevention:} Always use short table aliases (e.g., \texttt{t1}, \texttt{t2}) and explicitly qualify all column names with their respective table aliases, especially when \texttt{JOIN}s are involved.
    \item \textbf{Robustness:} Consider edge cases such as handling \texttt{NULL} values appropriately and string matching case conventions where applicable.
    \item \textbf{Thinking Process:} Before generating the final SQL query, step-by-step analyze the tables needed, the \texttt{JOIN} conditions, and the filtering criteria. Enclose your reasoning strictly within \texttt{<think>} and \texttt{</think>} tags.
\end{itemize}

\vspace{0.5em}
\noindent\textbf{Output Format:}\\
Please strictly follow this exact output structure. Do not output any other text or markdown outside of these tags:
\begin{Verbatim}[xleftmargin=1em, commandchars=\\\{\}]
<think>
Your step-by-step reasoning here.
</think>
<sql>
SELECT ...
</sql>
\end{Verbatim}
\end{promptbox}

\vspace{1em}

\begin{promptbox}{User Prompt --- Round 1}
\textbf{Database Schema:}\\
\texttt{\{db\_details\}}\\
\textit{This schema describes the database's structure, including tables, columns, primary keys, foreign keys, and any relevant relationships or constraints.}

\vspace{0.5em}
\noindent\textbf{Question:}\\
\texttt{\{question\}}
\end{promptbox}

\vspace{1em}

\begin{feedbackbox}{User Prompt --- Round 2+ (ODT Feedback Injection)}
Feedback: your previous SQL has structural issues: \texttt{\{error\_codes\}}, \\
\texttt{\{error\_descriptions\}} .

\vspace{0.5em}
\noindent Revise your previous answer using these issues. Revise your previous answer for SQLite. Put step-by-step reasoning inside \texttt{<think>...</think>}, then output exactly one corrected SQLite query inside \texttt{<sql>...</sql>}. Do not output any text outside those tags.
\end{feedbackbox}

\vspace{1em}
\noindent
\textit{Note:} When the ODT service fails to parse the predicted SQL, or when the gold SQL cannot be fully parsed and no cached structural profile is available, we inject a generic fallback message instead of clause-level diagnostic feedback.

\section{Case Study: Progressive Refinement with Structural Feedback}
\label{app:case_study_aggregation}

To qualitatively demonstrate the efficacy of our ODT-based dense feedback, we present a complete three-turn debugging trajectory. This case highlights how clause-level feedback helps the model correct aggregation granularity, projection targets, and redundant join structures during iterative SQL refinement.

\medskip\noindent\textbf{Question:}
\begin{quote}\itshape
Highly-rated suppliers refer to those who have received a rating of 4 or higher from customers. What are the names and average prices of contact lenses for highly-rated suppliers who have sold at least one lens?
\end{quote}

\begin{goldbox}{Gold SQL}
\begin{Verbatim}
SELECT s.supplier_name, AVG(cl.price) AS average_price 
FROM contact_lenses cl
JOIN suppliers s ON s.supplier_id = cl.supplier_id 
JOIN feedback f ON cl.lens_id = f.lens_id 
WHERE f.rating >= 4 
GROUP BY s.supplier_name 
HAVING COUNT(cl.lens_id) > 0;
\end{Verbatim}
\end{goldbox}

\vspace{1.5em}
\noindent\textbf{\Large Debugging Trajectory}
\vspace{0.5em}\hrule\vspace{1em}

\noindent\textbf{\large Round 1} \hfill \exwrong

\vspace{0.3em}
\noindent\textbf{Analysis:} The agent fundamentally misunderstands the required granularity. It attempts to group by \texttt{lens\_name} instead of \texttt{supplier\_name}, and over-complicates the query by unnecessarily joining the \texttt{sales} and \texttt{order\_items} tables to check for sales.
\vspace{0.5em}

\begin{modelbox}{Model Output}
\textbf{\textsf{Chain-of-Thought (Extracted):}} \\
{\small\itshape
To answer this question, we need to identify contact lenses from suppliers who have received a rating of 4 or higher... I will join contact\_lenses with suppliers, feedback, sales, and order\_items. Finally, I will group by lens name and calculate the average price. [snip]
}

\vspace{0.5em}
\textbf{\textsf{Predicted SQL:}}
\begin{Verbatim}[commandchars=\\\{\}]
SELECT \textcolor{red}{\textbf{t1.lens\_name}}, AVG(t1.price) AS average_price
FROM contact_lenses t1
JOIN suppliers t2 ON t1.supplier_id = t2.supplier_id
JOIN feedback t3 ON t1.lens_id = t3.lens_id
\textcolor{red}{\textbf{JOIN order\_items t4 ON t1.lens\_id = t4.lens\_id}}
WHERE t3.rating >= 4
GROUP BY \textcolor{red}{\textbf{t1.lens\_name}}
HAVING \textcolor{red}{\textbf{COUNT(DISTINCT t4.order\_item\_id) > 0}};
\end{Verbatim}

\end{modelbox}

\begin{feedbackbox}{ODT Feedback}
\begin{Verbatim}
Feedback: your previous SQL has structural issues: JOIN_KEY_MISMATCH,
FROM_OR_JOIN_TABLE_MISMATCH, SELECT_ERROR, GROUP_BY_ERROR. [...]
SELECT projection differs from reference. GROUP BY columns are not aligned
with reference, and redundant join structures are introduced.
\end{Verbatim}
\end{feedbackbox}

\vspace{1.5em}

\noindent\textbf{\large Round 2} \hfill \exwrong

\vspace{0.3em}
\noindent\textbf{Analysis:} Guided by the dense feedback, the agent successfully corrects the projection (\texttt{SELECT}) and grouping (\texttt{GROUP BY}) targets to \texttt{supplier\_name}. However, it stubbornly retains the redundant \texttt{sales} and \texttt{order\_items} tables, leading to a severe table mismatch warning.
\vspace{0.5em}

\begin{modelbox}{Model Output}
\textbf{\textsf{Chain-of-Thought (Extracted):}} \\
{\small\itshape
Based on the feedback, there are several issues... The selected column and
grouping key should be changed to the supplier name. I will revise the query
accordingly, but I still keep the transaction-table joins to check whether a lens
has been sold. [snip]
}

\vspace{0.5em}
\textbf{\textsf{Predicted SQL:}}
\begin{Verbatim}[commandchars=\\\{\}]
SELECT s.supplier_name, AVG(cl.price) AS average_price
FROM contact_lenses cl
JOIN suppliers s ON cl.supplier_id = s.supplier_id
JOIN feedback f ON cl.lens_id = f.lens_id
\textcolor{red}{\textbf{JOIN order\_items oi ON cl.lens\_id = oi.lens\_id}}
\textcolor{red}{\textbf{JOIN sales sa ON oi.order\_item\_id = sa.order\_id}}
WHERE f.rating >= 4
GROUP BY s.supplier_name;
\end{Verbatim}
\end{modelbox}

\begin{feedbackbox}{ODT Feedback}
\begin{Verbatim}
Feedback: your previous SQL has structural issues: JOIN_KEY_MISMATCH,
FROM_OR_JOIN_TABLE_MISMATCH. [...] FROM/JOIN tables differ from expected, and
the query introduces redundant transaction-table joins.
\end{Verbatim}
\end{feedbackbox}

\vspace{1.5em}

\noindent\textbf{\large Round 3} \hfill \exmatch

\vspace{0.3em}
\noindent\textbf{Analysis:} Acknowledging the specific
\texttt{FROM\_OR\_JOIN\_TABLE\_MISMATCH} error, the agent finally drops the
hallucinated transaction tables. It simplifies the query to the core tables and
recovers the correct aggregation granularity with
\texttt{GROUP BY s.supplier\_name} and \texttt{HAVING COUNT(cl.lens\_id) > 0},
achieving an exact execution match.
\vspace{0.5em}

\begin{modelbox}{Model Output}
\textbf{\textsf{Chain-of-Thought (Extracted):}} \\
{\small\itshape
Based on the new feedback... FROM\_OR\_JOIN\_TABLE\_MISMATCH: The FROM and JOIN clauses in the previous query included tables that are not necessary. I will only join contact\_lenses with suppliers and feedback. Then group by supplier name and calculate the average price. [snip]
}

\vspace{0.5em}
\textbf{\textsf{Predicted SQL:}}
\begin{Verbatim}
SELECT s.supplier_name, AVG(cl.price) AS average_price
FROM contact_lenses cl
JOIN suppliers s ON cl.supplier_id = s.supplier_id
JOIN feedback f ON cl.lens_id = f.lens_id
WHERE f.rating >= 4
GROUP BY s.supplier_name
HAVING COUNT(cl.lens_id) > 0;
\end{Verbatim}
\end{modelbox}

\end{document}